\documentclass{article}

\usepackage[preprint]{neurips_2026}

\usepackage[utf8]{inputenc} 
\usepackage[T1]{fontenc}    
\usepackage{hyperref}       
\usepackage{url}            
\usepackage{booktabs}       
\usepackage{amsfonts}       
\usepackage{amsmath}
\usepackage{amssymb}
\usepackage{nicefrac}       
\usepackage{microtype}      
\usepackage[table]{xcolor} 
\usepackage{enumitem}
\usepackage{graphicx}
\usepackage{multirow}
\usepackage{adjustbox}
\usepackage{array}
\usepackage{tikz}

\definecolor{QAFill}{HTML}{E8D4D4}
\definecolor{MathFill}{HTML}{D7E1EA}
\definecolor{CodingFill}{HTML}{EEE4C8}
\definecolor{tracegreen}{HTML}{D9F2D9}
\definecolor{tracered}{HTML}{F7C6C7}

\DeclareUnicodeCharacter{00B7}{\textperiodcentered}
\DeclareUnicodeCharacter{21B5}{\ensuremath{\hookleftarrow}}

\title{The Path Matters: Learning a Token-Commitment Policy for Diffusion Language Models}

\author{
Bohang Sun$^{1}$
\quad Max Zhu$^{1}$
\quad Francesco Caso$^{1}$
\quad Jindong Gu$^{2}$ \\
\textbf{Junchi Yu$^{2}$}
\quad \textbf{Philip Torr$^{2}$}
\quad \textbf{Pietro Li\`o$^{1}$}
\quad \textbf{Jialin Yu$^{2}$}\thanks{Corresponding author: \texttt{yu.jialin@outlook.com}.} \\
$^{1}$Department of Computer Science and Technology, University of Cambridge \\
$^{2}$Department of Engineering Science, University of Oxford
}

\begin{document}

\maketitle

\begin{abstract}
Diffusion large language models promise faster generation by refining many token positions in parallel, but this parallelism introduces a hidden control problem: which proposed tokens should be transferred into the partially decoded sequence at each step? We refer to this decision as token commitment. Existing frozen-generator decoders largely rely on hand-designed confidence rules or block-specific acceptance filters. We argue that token commitment can instead be learned as a reusable trace-state policy. We introduce \textsc{TraceLock}, a lightweight plug-in controller that instantiates this policy for a frozen diffusion language model. Since oracle commitment times are unavailable, \textsc{TraceLock} derives self-supervision from future stability: at decoding step \(t\), a proposed token for position \(i\) is labeled stable if it matches the final token at position \(i\) after the full decoding trace completes. The controller scores variable-length trace states and decides which active token proposals should be committed to the partially decoded sequence. Once trained for a given frozen backbone, the controller can be deployed across local-window widths, generation lengths, and step budgets without retraining or per-setting calibration. Experiments on question answering, mathematical reasoning, and code generation show that \textsc{TraceLock} improves the quality--step tradeoff over heuristic and learned baselines, with particularly stable behavior under cross-setting deployment. Diagnostic analyses show that its decisions are not reducible to scalar confidence, suggesting that frozen diffusion language models expose a learnable space of commitment trajectories beyond confidence-based decoding\footnote{Our code is released at \url{https://github.com/BobSun98/TraceLock}.}.
\end{abstract}

\section{Introduction}
Diffusion large language models (D-LLMs) have emerged as a promising alternative to autoregressive large language models (AR-LLMs), especially in masked discrete diffusion settings \citep{sahoo2024simple, ou2024your} such as LLaDA and Dream \citep{llada2025, ye2025dream}. Unlike AR-LLMs, which irreversibly append tokens from left to right, masked D-LLMs iteratively refine a partially masked sequence and can update multiple token positions in parallel. This parallel refinement avoids the fixed generation order of autoregressive decoding and creates the possibility of faster generation. In practice, however, high-quality D-LLM generation often still requires many denoising iterations, each involving bidirectional attention over the current sequence. Fast decoding is therefore a central challenge for deploying D-LLMs effectively \citep{fastdllm2025, chen2025dparallel, israel2025accelerating}. 

\begin{figure}[t]
\centering
\includegraphics[width=\linewidth]{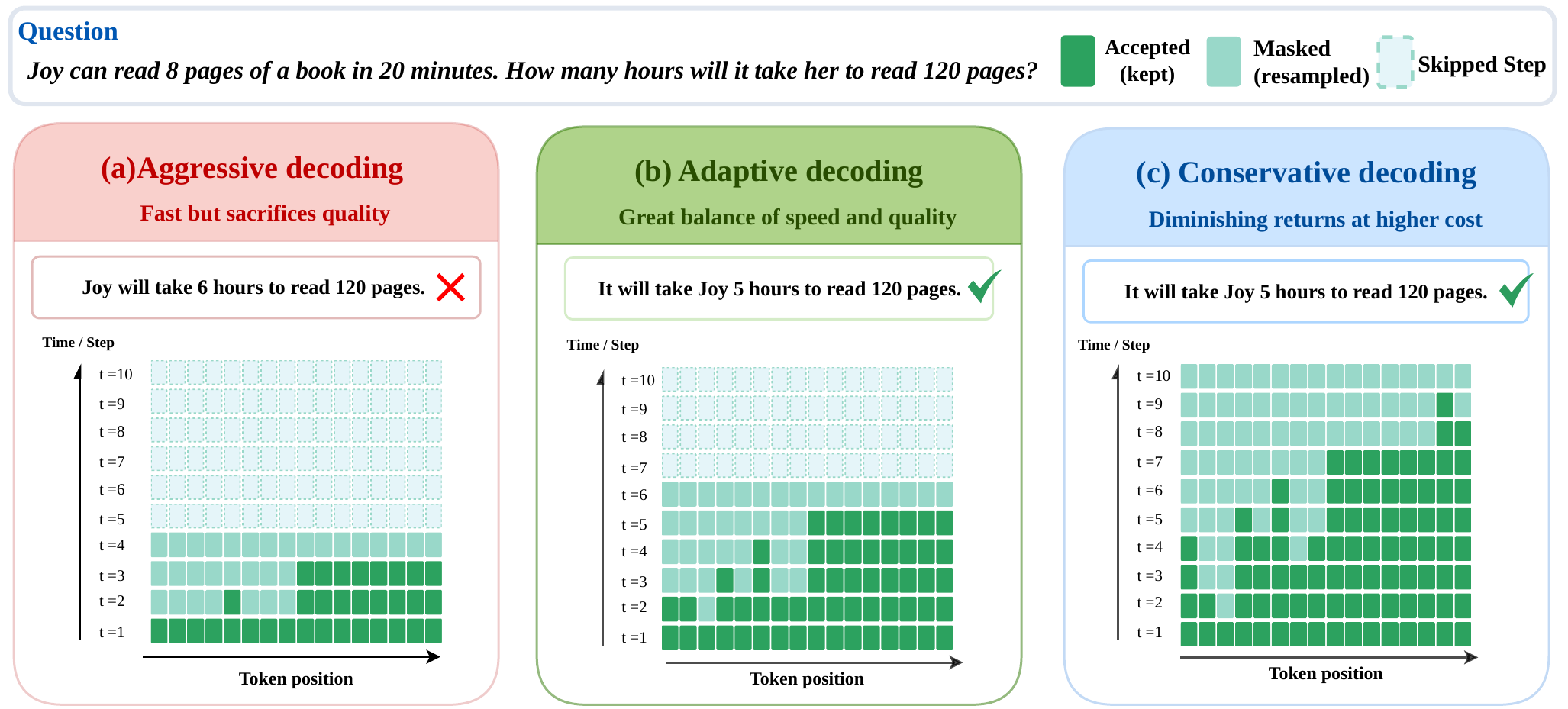}
\caption{Token commitment as a trace-selection problem in masked diffusion decoding. Aggressive decoding can be fast but may lock an incorrect trajectory, while conservative decoding keeps tokens revisable for longer at higher cost. The useful operating point is not a single global cutoff: it can vary across samples and across steps as the partial trajectory evolves.}
\label{fig:commitment_tradeoff}
\vspace{-5mm}
\end{figure}

Fast decoding for D-LLMs is often described as reducing the number or cost of denoising steps. In the frozen-generator setting, however, the generator is fixed and the main algorithmic freedom lies elsewhere: the decoder must decide which proposed tokens should stop being revised. \textit{This turns fast decoding into a token-commitment problem}. At each denoising step, the frozen model proposes token values for multiple positions, while the decoder must decide which positions should be committed and which should remain revisable. Choosing an effective commitment policy is non-trivial: committing too late limits speedup, while committing too early can lock in errors. Figure~\ref{fig:commitment_tradeoff} illustrates this tradeoff. Existing frozen-generator decoders can be understood as setting-specific ways of calibrating commitment. Training-free methods use hand-designed confidence thresholds, transfer rules, or block schedules \citep{fastdllm2025, fastdllmv22025, dong2025saber}, while learned filters such as Learn2PD learn an acceptance rule tied to a fixed block-level decoding interface \citep{bao2025learning}. This raises a natural question: \textit{can token commitment instead be learned as a reusable trace-state policy that can be deployed across local-window widths, generation lengths, and step budgets}?

Learning such a policy is difficult because the optimal commitment decisions are not directly observable. A completed generation tells us the final sequence, but not when each token should have stopped being revised. We therefore turn commitment learning into a self-supervised trace-prediction problem derived from completed diffusion traces. An intermediate token proposal is labeled as \emph{future-stable} if it matches the corresponding token in the final completed sequence. This future-stability target requires no human annotation or task-level correctness label. Although it is only a trace-relative signal, it provides a dense proxy for the online question of which active tokens are safe to commit. We propose \textsc{TraceLock}, a lightweight plug-in controller that learns a reusable commitment policy from these labels. The base D-LLM remains frozen and \textsc{TraceLock} only decides which active generation positions should remain revisable and which should become locked. Rather than designing or learning a setting-specific acceptance calibration, \textsc{TraceLock} scores variable-length trace states using frozen-model hidden states, short-range hidden-state dynamics, and prompt/active/locked context, with a single shared contextual scorer. We evaluate \textsc{TraceLock} on question answering, mathematical reasoning, and code generation tasks under multiple generation lengths and local-window regimes. The results show that trace-supervised contextual commitment improves the quality--step tradeoff over heuristic and learned decoding baselines in several settings.

\paragraph{Contributions.}
We make the following contributions: \textbf{Problem formulation.} We formulate efficient frozen-generator D-LLM decoding as a token-commitment problem, where the decoder decides when proposed tokens should become irreversible. \textbf{Algorithm.} We show that this commitment policy can be learned from completed diffusion traces using self-supervised future-stability labels, and instantiate it as \textsc{TraceLock}, an end-to-end learned commitment controller rather than a hand-designed or block-specific acceptance rule. \textbf{Empirical evidence.} Across mathematical reasoning, question answering, and code generation, we show that \textsc{TraceLock} improves the quality--step tradeoff over heuristic and learned baselines, remains stable across changes in local-window width and generation length, and learns commitment behavior beyond scalar confidence filtering.

\section{Related Work}
\label{sec:related}

\paragraph{D-LLM decoding acceleration.}

Discrete and masked diffusion language models have become an increasingly important alternative to autoregressive language modeling \citep{sahoo2024simple, ou2024your, llada2025, ye2025dream}. 
Existing acceleration methods intervene at different levels of the generation pipeline. 
Systems methods reduce the cost of individual denoising iterations, for example through caching or execution optimizations \citep{fastdllm2025, jiang2025d}. 
Model-adaptation methods modify the generator or its training objective so that faster or more aggressive parallel decoding becomes reliable, for example through finetuning, distillation, approximate joint sampling, or learned parallel decoding behavior \citep{chen2025dparallel, bansal2025enabling, israel2025accelerating}. 
Other methods change the generation process itself through hybrid, blockwise, or forcing-style formulations \citep{wang2025diffusion, arriola2025block}. 
These directions are complementary to our work: \textsc{TraceLock} does not train a new diffusion backbone or reduce the cost of an individual denoising step, but learns the frozen-generator inference-time policy that decides which proposed tokens should stop being revised.

\paragraph{Frozen-generator decoding acceleration.}
Our work is closest to methods that keep the D-LLM frozen and accelerate decoding by changing the token acceptance, remasking, or transfer policy. Training-free decoders such as Fast-dLLM and its follow-up variants use confidence-thresholded transfer, blockwise schedules, or related hand-designed rules to decide which positions should be revealed or finalized \citep{fastdllm2025, fastdllmv22025}. A broader line of work studies alternative heuristics for token ordering, confidence calibration, temporal modeling, or remasking behavior \citep{li2025diffusion, wang2025time, kim2025train, hong2025wide, dong2025saber}. These methods demonstrate that the decoding policy strongly affects the quality--efficiency tradeoff, but their acceptance boundaries are typically hand-designed and may require regime-specific calibration. Recently, Learn2PD \citep{bao2025learning} proposed a method that keeps the base D-LLM frozen and uses agreement with the final decoded sequence to train a lightweight token filter. \textsc{TraceLock} shares the idea of learning from future agreement, but differs in the policy being learned. Rather than learning a fixed block-level filter tied to a particular decoding interface, \textsc{TraceLock} learns a variable-length trace-state commitment policy. Our policy uses contextual hidden states, short-range state dynamics, and a sequence-conditioned threshold, enabling it to generalize across local windows, generation lengths, and step budgets without changing the architecture or checkpoint.

\section{Method}
\label{sec:method}

\begin{figure*}[t]
\centering
\includegraphics[width=0.98\textwidth]{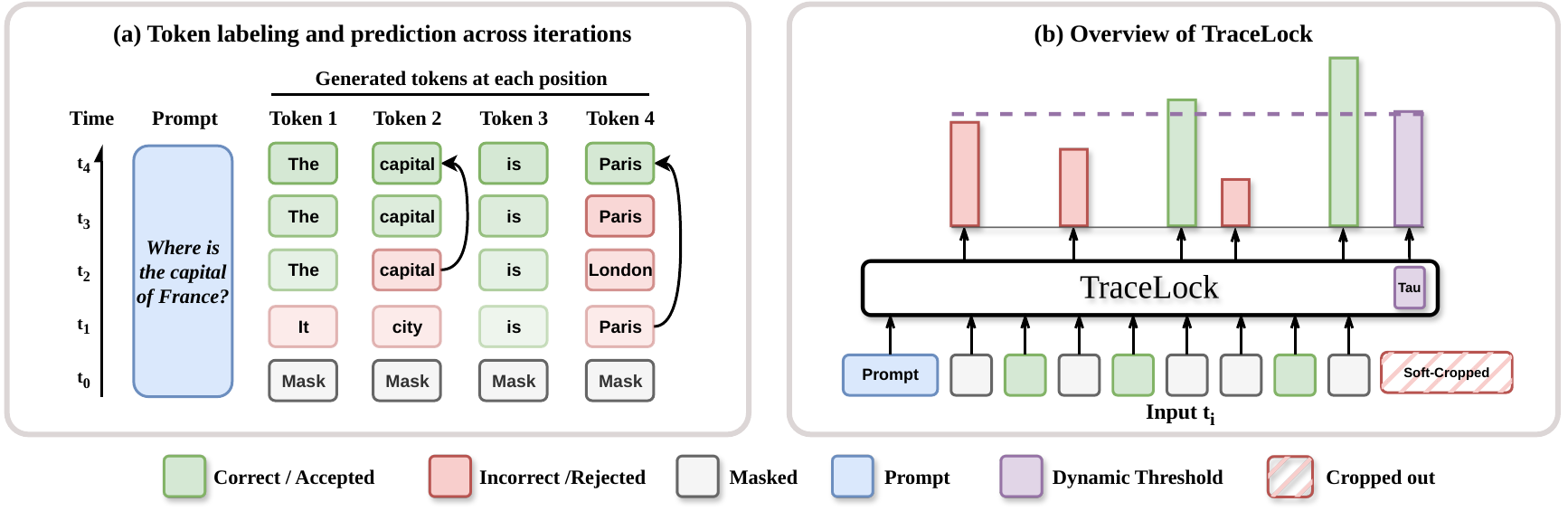}
\caption{Overview of \textsc{TraceLock}. (a) Completed diffusion traces provide dense token-level supervision: an intermediate token is labeled correct if it matches the token at the same position in the final completed trace. At deployment, \textsc{TraceLock} predicts the same future-stability event to decide whether a proposed token should be accepted. Here ``correct'' refers to agreement with the final trace token rather than task-level answer correctness. (b) \textsc{TraceLock} scores active tokens from the current trace state and compares them with a sequence-conditioned dynamic threshold. Soft-cropping restricts commitment decisions to the active local window, while prompt tokens and previously accepted tokens remain as context.}
\label{fig:tracelock_overview}
\end{figure*}

We describe \textsc{TraceLock} as a plug-in controller that implements a learned token-commitment policy inside the decoding loop of a frozen D-LLM. The base D-LLM proposes tokens and hidden states; \textsc{TraceLock} decides which proposed tokens should become final. At the control level, the goal is to learn a policy that maps the current trace state to commit-or-revise decisions. In the training process, completed generation traces provide self-supervised future-stability labels. For deployment, the same learned controller applies this policy to the current trace state without retraining or per-setting calibration. Figure~\ref{fig:tracelock_overview}(a) shows how completed traces define the future-stability labels used for training. Figure~\ref{fig:tracelock_overview}(b) shows the corresponding online decision at deployment: given the current partial trace, the controller predicts which active tokens are stable enough to lock.


\subsection{Problem Formulation}
\label{sec:method_problem}

Let a prompt occupy positions $1,\ldots,P$ and let the generation region have length $N$, so the total sequence length is $L=P+N$. A masked diffusion language model iteratively updates a sequence
\[
x_t \in \mathcal{V}^{L}, \qquad t = 0,\ldots,T,
\]
where $T$ is the total number of steps, $x_t$ are the predicted tokens at step $t$ and $\mathcal{V}$ is the vocabulary. Unfilled generation positions contain a special mask token. 

At each step, the frozen D-LLM with parameters $\psi$ produces token logits and hidden states
\[
(\ell_t, H_t) = F_\psi(x_t),
\]
where $\ell_{t,i}\in \mathbb{R}^{|\mathcal{V}|}$ is the token logit vector at position $i$ and $H_t$ is the corresponding internal hidden representations. Each position has a state represented by
\[
s_{t,i} \in \{\texttt{prompt}, \texttt{gen}, \texttt{locked}, \texttt{eot}\}.
\]
\texttt{Prompt} positions are immutable. \texttt{Locked} positions are generated tokens that have already been accepted and will not be revised. Active \texttt{gen} positions are revisable generation positions on which the controller acts. The \texttt{eot} state marks generated end-of-text or padding positions that terminate the decoded answer and are removed by the tokenizer when forming the final response. The central decision is therefore a transition 
\[
\texttt{gen} \rightarrow \texttt{locked}.
\]

Given the current candidate token
\[
\hat{x}_{t,i} = \arg\max_{v\in\mathcal{V}} \ell_{t,i,v},
\]
the controller decides whether to commit that candidate:
\[
u_{t,i} \in \{0,1\}, \qquad i \in \mathcal{G}_t,
\]
where $\mathcal{G}_t=\{i:s_{t,i}=\texttt{gen}\}$. If $u_{t,i}=1$, the position is filled with $\hat{x}_{t,i}$ and becomes \texttt{locked}; otherwise it remains masked for later refinement.

This formulation separates token proposal from token commitment. The D-LLM proposes candidate tokens; \textsc{TraceLock} only controls when candidates should stop being revised. The space of possible commitment traces is combinatorial: committing N positions across multiple non-empty rounds yields exponentially many possibilities. Exhaustively searching this trace space is infeasible, so our goal is to amortize the search by learning a policy that favors commitments already consistent with the eventual completed trace. \textsc{TraceLock} does not enlarge the expressive capacity of the base model; it biases decoding toward more favorable trajectories within the base model's reachable trace space.

\subsection{Learning a Commitment Policy from Future Stability}
\label{sec:method_learning}

\paragraph{Self-supervised future-stability labels.}
For each completed trace, we observe the final sequence $x^\star=x_T$ and the candidate token $\hat{x}_{t,i}$ proposed by the D-LLM at intermediate steps. We define a dense future label
\[
y_{t,i} = \mathbb{I}\left[\hat{x}_{t,i}=x_i^\star\right].
\]
This label asks whether the current candidate has already reached its final trace value. It is not a gold-answer label or a reward-model label; it is a self-supervised trace signal derived from the model's own completed rollout. The target is weak because a token can temporarily match the final value and later change, but it is dense, cheap, and aligned with the deployment-time question: which active positions are safe to lock now?

\paragraph{Trace collection and filtering.}
To generate training traces, we run the frozen base model, record candidate tokens and intermediate hidden representations at selected steps, and backfill labels by comparing each recorded step against the final completed sequence. Trace quality matters: low-quality traces would produce a poor controller since the controller is trained to reproduce the stability patterns in these traces. We therefore generate the main supervised trace pool with relatively fine block schedules, which preserves quality at the cost of slow inference. This lets \textsc{TraceLock} learn from higher-quality traces and later deploy under larger local windows or longer generation regimes.

Furthermore, we also filter traces before training. For open-ended text, we remove generations that are too short or exhibit obvious degeneration, including malformed, grammatically invalid, or highly repetitive text, using lightweight rule-based grammar checks when applicable \citep{language_tool_python}. For coding data, additional syntax and executable checks are used when available. 


\paragraph{Feature representation.}
The controller is driven by hidden representations from the frozen D-LLM. For each position $i$ at step $t$, let $\phi_{t,i}$ denote its trace-state feature which consists of an embedding of the last three hidden-states and differences between them, together with positional encodings. The intuition is that token stability is reflected not only in what the final hidden state is, but also in how the representation changes across late layers. Appendix Table~\ref{tab:feature_ablation} ablates these hidden-state components, and Table~\ref{tab:hidden_l2p_diagnostic} compares hidden-state inputs with confidence-only token filtering.

\paragraph{Contextual token scoring.}
\textsc{TraceLock} uses a lightweight Transformer encoder \citep{vaswani2017attention} over the trace-state feature sequence. Given $\phi_{t,1:L}$, it returns one raw stability logit per position,
\[
a_{t,i} = f_\theta(\phi_{t,1:L})_i.
\]
The controller is not parameterized as an MLP over a fixed-size block of confidence scores; it is a shared sequence policy over token trace states. Consequently, the number of active positions considered at deployment may vary with the local window or generation length, while the learned parameters remain unchanged. Contextual encoding lets each token decision depend on prompt tokens, already locked positions, and still-active generation positions in the current partial sequence.

\paragraph{Dynamic threshold.}
A model-level fixed threshold can be rigid: it may be too conservative for easy samples and too aggressive for difficult ones. \textsc{TraceLock} instead predicts a step-level threshold from the current trace state. We prepend a learned threshold token to the same feature sequence used for token scoring. After contextual encoding, a small threshold head applied to this token outputs a scalar threshold logit
\[
\tau_t = g_\theta(\phi_{t,1:L}).
\]
Here $g_\theta$ denotes the threshold branch of the \textsc{TraceLock} controller, not a separate language model. The decision logit for token $i$ is the margin
\[
m_{t,i} = a_{t,i} - \tau_t.
\]
Training this margin creates a direct competition between positive and negative examples: stable tokens push the threshold below their scores, while unstable tokens push it above. Thus the learned threshold acts as an adaptive caution value conditioned on the current partial sequence and the hidden states produced at this decoding step. Appendix Figure~\ref{fig:hidden_threshold} visualizes the learned threshold trajectories.

\paragraph{Training objective.}
The loss trains \textsc{TraceLock} to predict which tokens can be committed:
\begin{gather*}
    \mathcal{L}_{\text{dyn}}
        =
        \frac{1}{|\Omega|}
        \sum_{(t,i)\in\Omega} W(m_{t, i}, y_{t, i})
        \operatorname{BCE}(m_{t,i}, y_{t,i})
        +
        \lambda_\tau \tau_t^2 , \\
    W(m, y) =
    \begin{cases}
        1, & \text{if } (m>0) = y \\
        k, & \text{if } (m>0) \neq y.
    \end{cases}
\end{gather*}

where $\Omega$ runs over active generation positions excluding prompt and locked tokens,
\[
\Omega=\{(t,i):s_{t,i}=\texttt{gen}\}.
\]
The first term enforces correct predictions from the model, the second term encourages stability to ensure logits remain in a reasonable range. The weighting factor $W(m_{t, i}, y_{t, i})$ is used to strongly penalize incorrect stability decisions, since incorrect commitments strongly degrade the final prediction. We set $k=4$ in our experiments. 

During training, we simulate varying window sizes with random prefix cropping. A visible generation prefix is sampled as 
\[
K = \operatorname{round}(rN), \qquad r\sim \operatorname{Uniform}(0.25,1),
\]
and masks attention beyond $P+K$. This simulates the partial contexts seen by local-window deployment and helps the same controller generalize across different generation lengths and windows. 

\subsection{Deploying the Learned Policy with \textsc{TraceLock}}
\label{sec:method_deployment}
At deployment, \textsc{TraceLock} uses a local soft-crop window rather than committing over the entire generation region at every step. Let $p_t=\min \mathcal{G}_t$ be the first still-active generation position and let $w$ be the window width. The controller acts on window 
\[
W_t = [p_t, \min(p_t+w-1,L)]
\]
where $w$ is the desired window length. Operationally, this is a two-pointer policy: a slow pointer tracks the first unresolved position, while the fast right pointer defines the active decision window. This avoids strict divisibility constraints from fixed block decoding and keeps the candidate set from shrinking too sharply near block boundaries.

Within the active window, \textsc{TraceLock} converts the learned margin into a commit probability and commits positions satisfying $ \operatorname{sigmoid}(m_{t,i}) > \tilde{\tau}$, where $\tilde{\tau}$ is a fixed global operating threshold rather than a per-setting calibration parameter. We set $\tilde{\tau}=0.95$ for all main experiments, making the controller conservative against incorrect early commitments. If no position passes the threshold, the policy commits the highest-scoring active position as a fallback. This prevents the decoding loop from stalling while still allowing high-stability tokens to be accepted early. Because the controller scores the currently exposed trace state rather than a fixed block interface, the same checkpoint can be evaluated under different window widths, generation lengths, and step budgets. Table~\ref{tab:l2p_generalization} and Figure~\ref{fig:tracelock_genlen_window_sweep} evaluate this deployment flexibility.

\section{Experiments}
\label{sec:experiments}

We evaluate \textsc{TraceLock} as an inference-time commitment controller for frozen diffusion language models. The experiments test three claims: whether a learned trace-state policy improves the quality--step tradeoff over heuristic and learned baselines, whether the same controller remains usable across local-window and generation-length changes, and whether its decisions differ from confidence filtering. We evaluate these questions on \colorbox{MathFill}{mathematical reasoning}, \colorbox{QAFill}{question answering}, and \colorbox{CodingFill}{code generation}.

\subsection{Experimental Setup}

\paragraph{Backbones.}
All decoding policies are evaluated on frozen D-LLM backbones. We use LLaDA \citep{llada2025} and Dream \citep{ye2025dream} as the frozen backbones.

\paragraph{Tasks.}
Training traces are collected from the frozen backbone and then labeled by future stability, as described in Section~\ref{sec:method_learning}. The trace pool covers GSM8K for math \citep{cobbe2021gsm8k}, Natural Questions and Alpaca-style instruction data for QA \citep{kwiatkowski2019natural, alpaca2023}, and KodCode for coding \citep{xu2025kodcode}.\footnote{We release the processed data assets used for the Natural Questions and KodCode subsets at \url{https://huggingface.co/datasets/BOB12311/natural-questions-slim-short-answer} and \url{https://huggingface.co/datasets/BOB12311/kodcode-humaneval-like}.} Evaluation uses GSM8K final-answer accuracy for math, Alpaca-style QA prompts for answer ranking, and HumanEval execution tests for code generation \citep{chen2021humaneval}. 

\paragraph{Baselines.}
We compare \textsc{TraceLock} with policies that use the same frozen backbone: random transfer, confidence transfer, Fast-dLLM-style thresholded decoding \citep{fastdllm2025, fastdllmv22025}, and Learn2PD (L2P) \citep{bao2025learning}. For Learn2PD, we train a separate model for each block size. The default \textsc{TraceLock} deployment uses the local soft-crop window and sequence-conditioned dynamic threshold from Section~\ref{sec:method_deployment}. Appendix Table~\ref{tab:tracelock_variants} reports optional self-training and RL adaptation variants.

\paragraph{Metrics.}
For math, a response is correct if its extracted final answer matches the GSM8K target. For coding, we report HumanEval Pass@1; syntax errors, runtime errors, and timeouts are counted as failures. For QA, we anonymize the candidate answers for the same prompt, rank them in a single batch with Llama-3.1-8B-Instruct \citep{grattafiori2024llama3} under a shared rubric, and report average rank, where lower is better. All QA average-rank results in the paper use this within-prompt batch-ranking protocol,  but ranks are comparable only within the candidate set used by each table. Efficiency is measured by average executed decoding steps. 

\subsection{Main Results}
Table~\ref{fig:main_result_tradeoff} summarizes the main quality--step tradeoff. The left panel shows a compact LLaDA slice at generation length / block size $128/32$ and $256/64$, while the right panel reports six model--task averages: for each backbone and task, primary metrics are averaged over the tested length/window-or-block size settings and steps are first normalized by the maximum scheduled steps for that generation length before averaging. Full per-setting numbers for both LLaDA and Dream are reported in Appendix Table~\ref{tab:main_results}.

Overall, \textsc{TraceLock} improves the quality--step frontier, but its operating point differs across backbones. On Dream, it often improves both task score and decoding steps relative to the strongest baselines. On LLaDA, it typically prioritizes final quality, achieving the best or near-best scores while using more decoding steps than the most aggressive baselines. These results support the view that learned trace-state commitment provides a useful path-selection mechanism for frozen D-LLM decoding.

\paragraph{Other Variations}

Beyond the default frozen-controller setting (\textsc{TraceLock}), we also explored two optional extensions: (1) target distribution adaptation (\textsc{TraceLock-ST}) and (2) reinforcement-learning-based refinement (\textsc{TraceLock-RL}). These extensions are not required for the main results, but suggest that the learned commitment policy can be further improved with additional adaptation (Appendix \ref{appendix:other-variations} and Table \ref{tab:tracelock_variants}). \textsc{TraceLock-ST} mainly reduces the number of denoising steps, while \textsc{TraceLock-RL} can further improve final task performance in selected settings.

\begin{table*}[t]

\centering
\newlength{\mainresultpanelheight}
\setlength{\mainresultpanelheight}{0.45\textwidth}
\begin{minipage}[t][\mainresultpanelheight][t]{0.39\textwidth}
\centering
\vspace{0.01\textwidth}
\footnotesize
\setlength{\tabcolsep}{2.4pt}
\renewcommand{\arraystretch}{1.3}
\resizebox{\linewidth}{!}{%
\begin{tabular}{lccccc}
\toprule
Set & Rand. & Conf. & Fast & L2P & TL \\
\midrule
\rowcolor{MathFill}\multicolumn{6}{c}{\textbf{Math} (ACC $\uparrow$)} \\
\rowcolor{MathFill}128/32 & 67.1 & 74.4 & 74.8 & 74.8 & \textbf{74.9} \\
\rowcolor{MathFill}256/64 & 69.1 & 79.8 & 79.4 & 78.3 & \textbf{80.0} \\
\midrule
\rowcolor{QAFill}\multicolumn{6}{c}{\textbf{QA} (Avg.\ rank $\downarrow$)} \\
\rowcolor{QAFill}128/32 & 3.42 & 2.89 & 3.06 & 3.02 & \textbf{2.60} \\
\rowcolor{QAFill}256/64 & 3.51 & 2.74 & 2.87 & 3.21 & \textbf{2.68} \\
\midrule
\rowcolor{CodingFill}\multicolumn{6}{c}{\textbf{Coding} (Pass@1 $\uparrow$)} \\
\rowcolor{CodingFill}128/32 & 14.0 & 31.1 & 29.9 & 28.7 & \textbf{32.3} \\
\rowcolor{CodingFill}256/64 & 15.2 & 36.0 & 35.4 & 32.3 & \textbf{37.2} \\
\bottomrule
\end{tabular}
}
\end{minipage}
\hfill
\begin{minipage}[t][\mainresultpanelheight][c]{0.6\textwidth}
\centering
\vspace{0pt}
\includegraphics[width=\textwidth,height=\mainresultpanelheight]{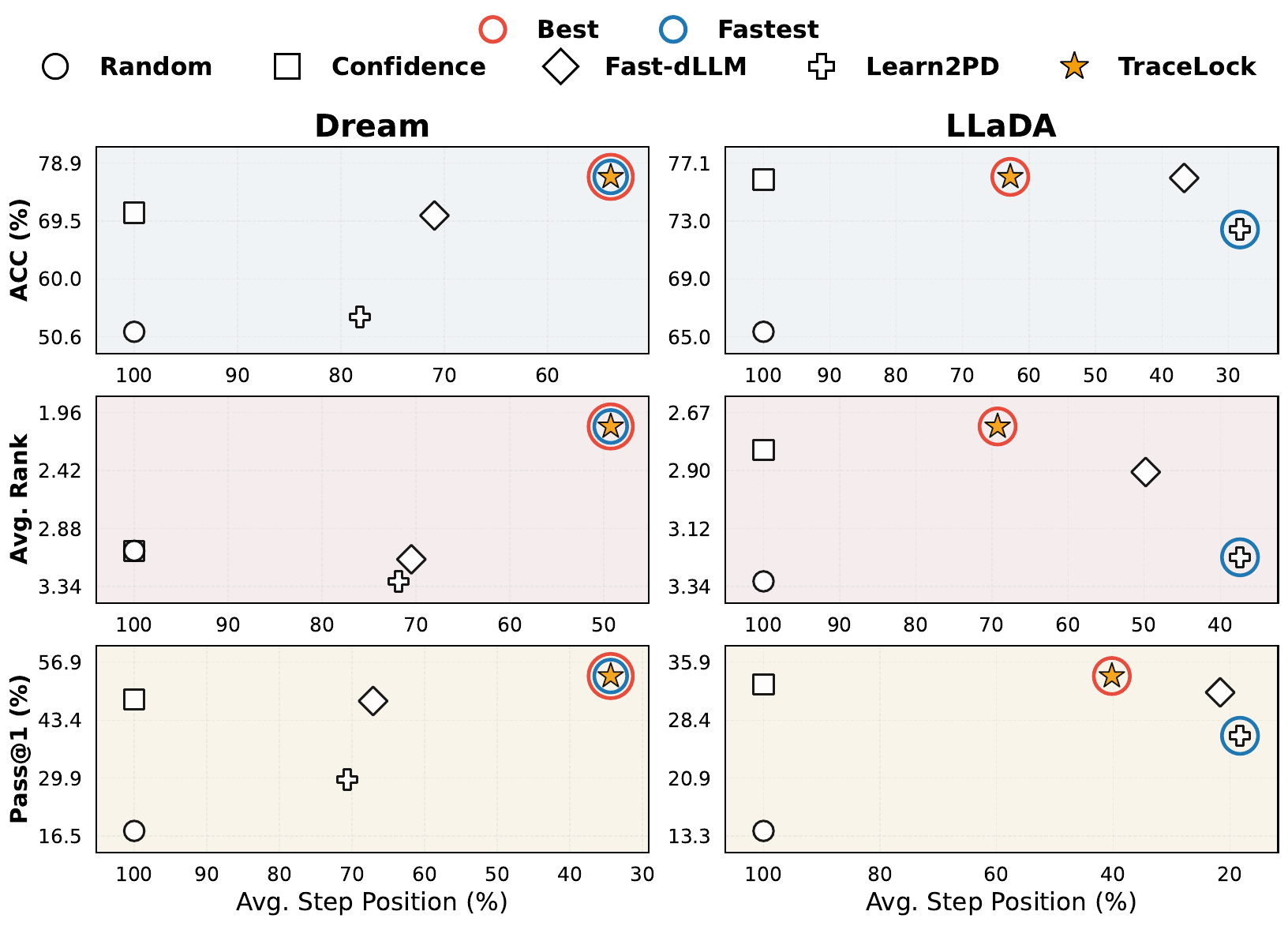}
\end{minipage}
\caption{Compact main-result summary and average quality--step tradeoff. Left: scores with LLaDA using generation length / window-or-block size $128/32$ and $256/64$ over the three tasks. The full table is available in Appendix Table~\ref{tab:main_results}. Right: plots of average scores versus normalized number of steps taken; fewer steps indicate faster sampling. Higher scores are better for Math and Coding, while lower scores are better for QA average rank.}
\label{fig:main_result_tradeoff}
\end{table*}

\subsection{Ablation Study}

\begin{figure*}[t]
\centering
\includegraphics[width=\textwidth]{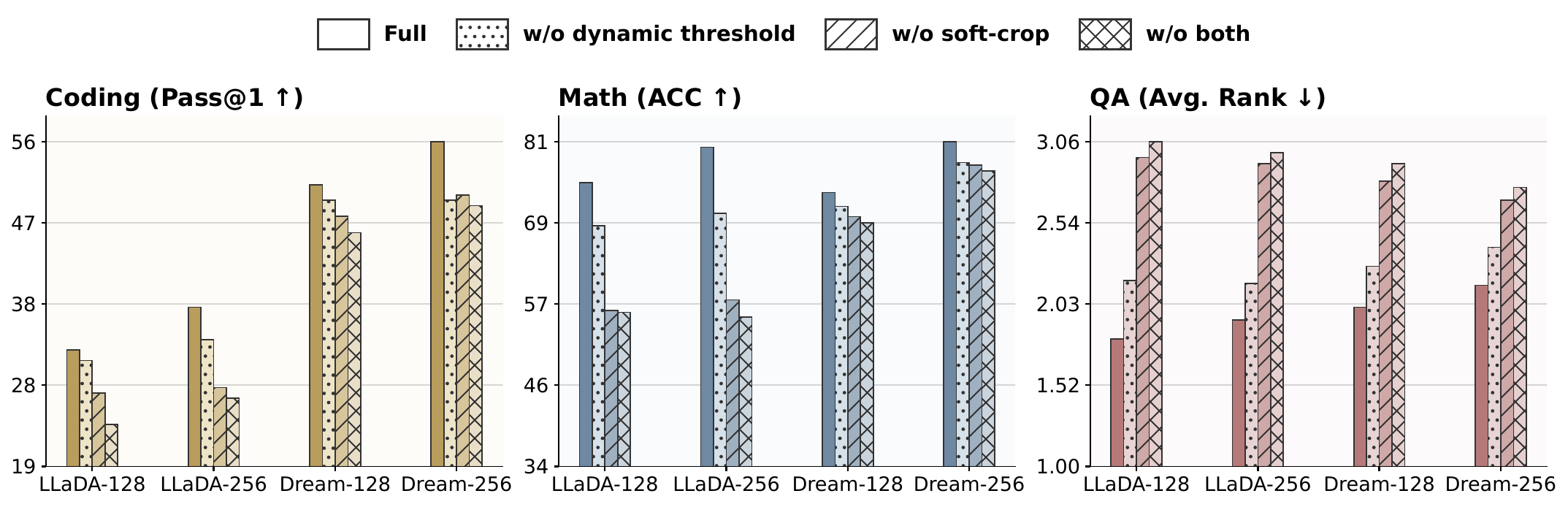}
\caption{Deployment-mechanism ablations by domain. Each panel shows the full \textsc{TraceLock} deployment, removal of soft-crop, replacement of the dynamic threshold with a static threshold, and removal of both mechanisms. Full numbers are in Appendix Table~\ref{tab:deployment_ablation}.}
\label{fig:deployment_ablation_panels}
\end{figure*}

\textsc{TraceLock}'s default deployment uses two implementation choices: soft-crop defines the active decision window, while the learned sequence-conditioned threshold provides a trace-conditioned acceptance boundary. Figure~\ref{fig:deployment_ablation_panels} ablates these mechanisms across coding, math, and QA. The full controller is consistently better than removing either choice, and removing both is the weakest setting overall. This indicates that the learned future-stability policy benefits from both a local decision scope and a trace-conditioned acceptance boundary.


\subsection{Comparison to Blockwise and Confidence-Based Commitment}

\paragraph{Goal of comparison.}
The closest alternatives to \textsc{TraceLock} are learned blockwise filters such as Learn2PD and 
training-free confidence-based transfer rules. Both decide which tokens should be accepted 
during diffusion decoding, but they induce different inductive biases. Learn2PD learns a 
filter tied to a fixed block interface, while confidence-based methods use scalar token 
confidence as the main acceptance signal. We therefore use this section to test two 
diagnostic questions: whether \textsc{TraceLock} remains stable when the deployment window changes, 
and whether its trajectories differ from confidence filtering on the same prompts.

\paragraph{Generalization beyond a fixed block interface.}
A practical D-LLM decoder should remain usable when the generation length or local-window size changes, since both are deployment choices determined by the task length and latency budget. Fixed blockwise filters can be brittle under such changes because their input-output interface is tied to a particular block shape. In contrast, \textsc{TraceLock} scores variable-length trace states and can therefore be deployed under different local windows without changing the policy architecture. Table~\ref{tab:l2p_generalization} compares Learn2PD and \textsc{TraceLock} when the test-time block or local window is enlarged. Across GSM8K, Alpaca-style QA, and HumanEval, Learn2PD often degrades substantially under these transfers, especially for longer generation lengths. \textsc{TraceLock} is consistently more stable, supporting the claim that it learns a reusable contextual commitment policy rather than a block-specific acceptance filter.

\begin{figure}[t]
\centering
\includegraphics[width=\linewidth]{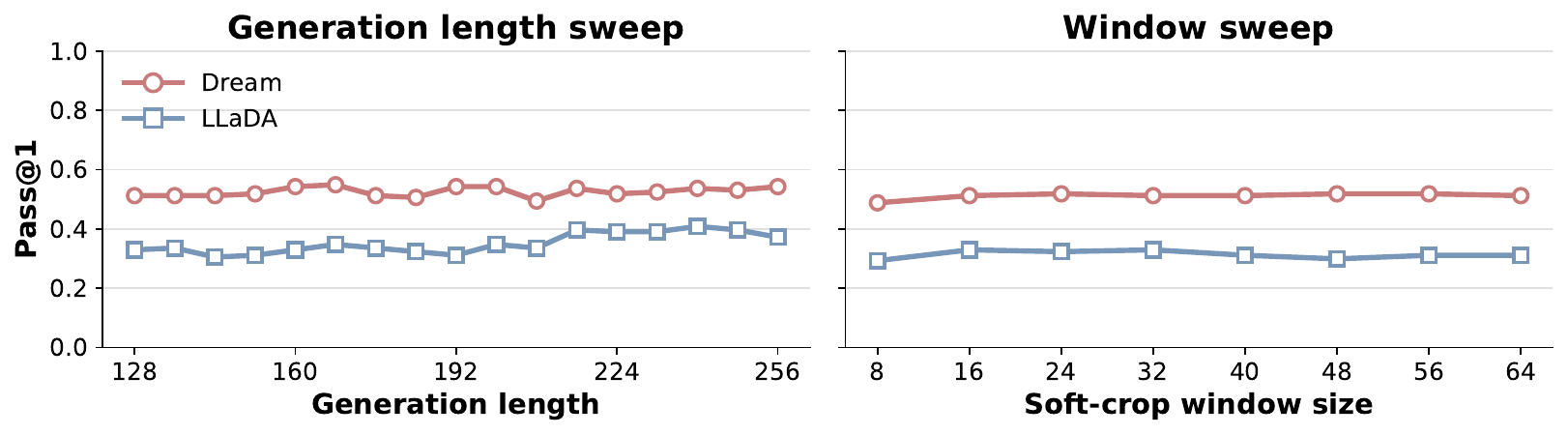}
\caption{Generation-length and pointer-window sweeps on HumanEval. The same trained \textsc{TraceLock} controller is used throughout, and performance remains stable across the tested range.}
\label{fig:tracelock_genlen_window_sweep}
\end{figure}

Figure~\ref{fig:tracelock_genlen_window_sweep} further sweeps \textsc{TraceLock} over generation lengths and window sizes without retraining. Performance remains broadly stable across both backbones, suggesting that the learned policy is not calibrated only to one deployment shape.

\begin{table}[t]
\centering
\small
\caption{Generalization from the original block/window to a larger one. GSM8K reports accuracy (\%), QA reports average rank under four-candidate within-prompt ranking, and HumanEval reports Pass@1 (\%). Parentheses show the change from original to generalized; for QA, lower rank is better, so positive rank changes indicate degradation.}
\label{tab:l2p_generalization}
\setlength{\tabcolsep}{4pt}
\renewcommand{\arraystretch}{1.08}
\begin{tabular}{lcccccc}
\toprule
Method & Gen len & Block change
& \multicolumn{2}{c}{LLaDA}
& \multicolumn{2}{c}{Dream} \\
\cmidrule(lr){4-5} \cmidrule(lr){6-7}
& & & {\scriptsize Original} & Generalized & {\scriptsize Original} & Generalized \\
\midrule
\rowcolor{MathFill}
Learn2PD & 128 & 32$\rightarrow$64 & {\scriptsize 74.8} & {\scriptsize 64.4} (-10.4 pp) & {\scriptsize 69.1} & {\scriptsize 66.8} (-2.3 pp) \\
\rowcolor{MathFill}
TraceLock & 128 & 32$\rightarrow$64 & {\scriptsize 74.9} & {\scriptsize 73.0} \textbf{(-1.9 pp)} & {\scriptsize 73.5} & {\scriptsize 71.5} \textbf{(-2.0 pp)} \\
\midrule
\rowcolor{MathFill}
Learn2PD & 256 & 64$\rightarrow$128 & {\scriptsize 57.4} & {\scriptsize 41.1} (-16.3 pp) & {\scriptsize 64.6} & {\scriptsize 45.2} (-19.4 pp) \\
\rowcolor{MathFill}
TraceLock & 256 & 64$\rightarrow$128 & {\scriptsize 80.0} & {\scriptsize 76.5} \textbf{(-3.5 pp)} & {\scriptsize 80.8} & {\scriptsize 80.7} \textbf{(-0.1 pp)} \\
\midrule
\rowcolor{QAFill}
Learn2PD & 128 & 32$\rightarrow$64 & {\scriptsize 2.28} & {\scriptsize 2.77} (+0.49 rank) & {\scriptsize 2.65} & {\scriptsize 3.14} (+0.49 rank) \\
\rowcolor{QAFill}
\textsc{TraceLock} & 128 & 32$\rightarrow$64 & {\scriptsize 2.07} & {\scriptsize 2.18} \textbf{(+0.12 rank)} & {\scriptsize 1.81} & {\scriptsize 2.05} \textbf{(+0.24 rank)} \\
\midrule
\rowcolor{QAFill}
Learn2PD & 256 & 64$\rightarrow$128 & {\scriptsize 2.33} & {\scriptsize 2.71} (+0.38 rank) & {\scriptsize 2.62} & {\scriptsize 3.15} (+0.53 rank) \\
\rowcolor{QAFill}
\textsc{TraceLock} & 256 & 64$\rightarrow$128 & {\scriptsize 2.03} & {\scriptsize 2.23} \textbf{(+0.20 rank)} & {\scriptsize 1.82} & {\scriptsize 2.14} \textbf{(+0.32 rank)} \\
\midrule
\rowcolor{CodingFill}
Learn2PD & 128 & 32$\rightarrow$64 & {\scriptsize 28.7} & {\scriptsize 25.0} (-3.7 pp) & {\scriptsize 48.8} & {\scriptsize 45.7} (-3.1 pp) \\
\rowcolor{CodingFill}
\textsc{TraceLock} & 128 & 32$\rightarrow$64 & {\scriptsize 32.3} & {\scriptsize 31.7} \textbf{(-0.6 pp)} & {\scriptsize 51.2} & {\scriptsize 51.2} \textbf{(+0.0 pp)} \\
\midrule
\rowcolor{CodingFill}
Learn2PD & 256 & 64$\rightarrow$128 & {\scriptsize 32.3} & {\scriptsize 23.2} (-9.1 pp) & {\scriptsize 31.7} & {\scriptsize 28.0} (-3.7 pp) \\
\rowcolor{CodingFill}
\textsc{TraceLock} & 256 & 64$\rightarrow$128 & {\scriptsize 37.2} & {\scriptsize 35.4} \textbf{(-1.8 pp)} & {\scriptsize 56.1} & {\scriptsize 56.1} \textbf{(+0.0 pp)} \\
\bottomrule
\end{tabular}
\end{table}

\paragraph{Additional comparisons to confidence-based policies.}

Confidence is still a natural reference point for any token-commitment policy, since it is the strongest training-free signal available from the base diffusion model. We therefore include several appendix diagnostics that compare \textsc{TraceLock} against confidence-based behavior beyond final task scores. Table~\ref{tab:trace_divergence_summary} measures trajectory-level divergence from confidence filtering on matched QA prompts; Figure~\ref{fig:score_confidence} compares \textsc{TraceLock} scores with token confidence over decoding time; Figures~\ref{fig:trace_compare1} and~\ref{fig:trace_compare2} visualize the two policies on matched qualitative traces; and Table~\ref{tab:hidden_l2p_diagnostic} replaces confidence inputs in a Learn2PD-style blockwise filter with hidden-state features. Together, these diagnostics show that \textsc{TraceLock} remains related to confidence but is not simply a different threshold on the same confidence ordering.


\section{Conclusion}

\textbf{Findings.} We presented \textsc{TraceLock}, a lightweight plug-in controller that implements a learned token-commitment policy for masked diffusion language models. For a frozen diffusion language model, fast decoding is not only a matter of reducing denoising steps, but also a trajectory-selection problem: the decoder must decide which token proposals to commit to the partially decoded sequence and which positions to leave masked for later refinement. We show that this commitment policy can be learned from completed decoding traces using future-stability labels as dense self-supervision over variable-length trace states. Across mathematical reasoning, question answering, and code generation, \textsc{TraceLock} improves the quality--step tradeoff over heuristic and learned decoding baselines. Its stability across local-window widths and generation lengths suggests that token commitment can be learned as a reusable trace-state policy rather than as a block-specific acceptance filter. Together, these results show that diffusion language model performance is shaped not only by the frozen generator, but also by the path chosen through its iterative refinement space.

\textbf{Limitations and scope.}
Future-stability supervision is relative to the model's own completed traces rather than absolute task correctness, so the learned policy can inherit biases from the trace distribution. In particular, the controller learns to predict agreement with high-quality completed rollouts, not whether a token is correct under an external task objective. Our experiments should therefore be interpreted as evidence for reusable commitment policies within a fixed frozen-generator family and related decoding interfaces, rather than as evidence for a universal controller across unrelated diffusion generators, sampling procedures, or task distributions. Finally, judge-based QA evaluation should be interpreted as complementary to automatically checkable metrics such as GSM8K accuracy and HumanEval execution.


\newpage

\bibliographystyle{plainnat}
\bibliography{neurips_2026}

\appendix
\section{Use of LLMs}

Large language models were used as research and writing assistants during the preparation of this work. Their use included drafting and editing paper text, organizing literature, checking exposition for clarity, assisting with code debugging, and helping summarize related work. The authors manually verified the scientific claims, citations, experimental numbers, and final text of this paper.

This assistance was separate from the core experimental pipeline. The proposed method, trace collection procedure, controller training, evaluation scripts, and reported results were designed and checked by the authors. Where LLM-based components are part of the research artifact itself, such as LLM-as-Judge comparisons for QA or reward-model scoring for the reinforcement-learning extension, they are treated as evaluation or optimization components and described in the corresponding method and experiment sections. These model-based evaluations are used as complements to task-specific metrics rather than as replacements for automatic verification where such verification is available.

\section{Detailed Related Work}


\subsection{Implemented baselines.}

\paragraph{Random and confidence transfer.}
We include the two native remasking strategies used by the LLaDA and Dream sampling code: random transfer and confidence transfer \citep{llada2025,ye2025dream}. Generation follows a semi-autoregressive (Semi-AR) schedule: the output region is divided into local blocks, and decoding proceeds block by block from left to right. At each denoising step, the frozen D-LLM proposes tokens for the currently masked positions in the active block. For stability, we transfer one position per step for both random and confidence transfer. Random transfer selects this position uniformly at random from the active block. Confidence transfer selects the most confident candidate in the active block. For LLaDA, confidence is the softmax probability of the proposed token; for Dream, we follow the native implementation and use its entropy-based confidence score.

\paragraph{Fast-dLLM.}
Fast-dLLM is a stronger confidence-based decoding baseline that uses thresholded token transfer \citep{fastdllm2025,fastdllmv22025}. At each step, candidate tokens whose confidence exceeds a fixed threshold are transferred immediately. If no candidate reaches the threshold, the policy falls back to transferring the highest-confidence token, preventing the decoding process from stalling. Following the public Fast-dLLM implementation, we set the threshold to $0.9$ in all experiments.

\paragraph{Learn2PD.}
We also include Learn2PD, a learned parallel-decoding filter for frozen D-LLMs \citep{bao2025learning}. Following the original implementation, we train a separate confidence filter for each backbone and block size in $\{\text{LLaDA}, \text{Dream}\}\times\{32,64,128\}$. To match the original Learn2PD setup, we use the FLAN instruction-tuning data for filter training \citep{wei2022finetunedlanguagemodelszeroshot,longpre2023flancollectiondesigningdata}. For each setting, we sample $100$ prompts per task, generate traces with the corresponding frozen backbone, and train a two-layer MLP filter on the resulting confidence traces. At inference time, we use the thresholds from the original codebase: $0.96$ for LLaDA and $0.9$ for Dream.

\subsection{Other related works}

\textbf{Prophet} asks whether a D-LLM already ``knows'' the answer before all scheduled denoising steps are complete and uses this observation for early commit or early stopping \citep{li2025diffusion}. It is highly relevant because it frames decoding as a commitment problem, but its main decision is sequence-level stopping rather than token-level contextual locking. We therefore treat it as a potential future baseline for early-exit behavior, not as a direct replacement for a token controller.

\textbf{APD} uses a small auxiliary autoregressive model to guide adaptive parallel decoding \citep{israel2025accelerating}. It is related because it learns an adaptive decoding behavior at inference time, but it changes the inference stack by adding a second generator. This makes it less direct for our frozen-D-LLM-only setting, where the goal is to learn from the hidden states and traces of the target D-LLM itself.

\textbf{DAPD} is a training-free dependency-aware method that uses attention to infer which masked tokens can be decoded in parallel \citep{kim2026dapd}. It is complementary to \textsc{TraceLock}: DAPD uses an explicit dependency graph, while \textsc{TraceLock} learns a stability score from trace states. We do not include it in the current experiments because our present baseline suite focuses on confidence and learned-controller comparisons; dependency-aware scheduling is a natural future addition.

\textbf{dParallel} adapts the generator through distillation so that the model becomes more suitable for parallel decoding \citep{chen2025dparallel}. This is an important generator-adaptation baseline, but it answers a different question from \textsc{TraceLock}. Our method asks what can be gained by learning an inference-time policy on top of a fixed model, whereas dParallel changes the model being decoded.

\textbf{RemeDi} adds remasking ability to the model itself through remask-aware supervised finetuning and reinforcement learning \citep{huang2025remedi}. It is closely related at the level of motivation because both methods care about premature commitment. The key difference is that RemeDi trains a D-LLM to revise tokens, while \textsc{TraceLock} leaves the generator frozen and controls when proposed tokens are locked.

\textbf{D2F} and \textbf{Block Diffusion} change the generation geometry by combining autoregressive and diffusion-style behavior \citep{wang2025diffusion, arriola2025block}. These methods can improve the speed--quality tradeoff, but they operate through a different backbone or decoding formulation. We therefore view them as complementary generator-design baselines rather than direct plug-in controller baselines.

\textbf{RCD} argues that confidence-based remasking wastes computation by discarding unresolved token representations, then recycles those residual representations into later denoising steps \citep{hu2026rcd}. This is conceptually aligned with our claim that trace state matters beyond local confidence. However, RCD modifies the model architecture or training pipeline, while \textsc{TraceLock} only learns an auxiliary commitment policy.

\textbf{I-DLM} introduces an introspective diffusion-language-model paradigm with verification-like acceptance during decoding \citep{yu2026idlm}. It is relevant because it explicitly studies acceptance and self-consistency, but it is not a plug-in policy for an existing frozen D-LLM. Comparing against it would require a different model family and serving stack.

Finally, systems and execution-level methods such as \textbf{Fast-dLLM v2}, \textbf{d$^2$ Cache}, and \textbf{DPad} reduce the cost of each denoising step through caching, suffix-token pruning, or attention-computation optimizations\citep{fastdllmv22025, jiang2025d, chen2025dpad}. These methods are orthogonal to \textsc{TraceLock}: a learned commitment policy can in principle be combined with lower-cost model execution. For this reason, our current experiments focus on decoding-policy comparisons rather than systems-level throughput engineering.

\section{Full Main Results}

Table~\ref{tab:main_results} reports the full per-setting numbers summarized in Table~\ref{fig:main_result_tradeoff}.

\begin{table*}[t]
\centering
\caption{Combined main results across math, QA, and coding. Within each domain block, each policy is reported with a primary task metric and the average number of executed decoding steps. Lower is better only for QA average rank. Appendix Table~\ref{tab:qa_appendix_top2_critic} gives additional QA top-two and critic metrics.}
\label{tab:main_results}
\setlength{\tabcolsep}{3pt}
\renewcommand{\arraystretch}{1.05}
\colorbox{MathFill}{%
\parbox{\dimexpr\textwidth-2\fboxsep\relax}{%
\centering
\resizebox{\textwidth}{!}{%
\begin{tabular}{lllcccccccccc}
\toprule
Model & Gen len & Win./Block & \multicolumn{2}{c}{Random} & \multicolumn{2}{c}{Confidence} & \multicolumn{2}{c}{Fast-dLLM} & \multicolumn{2}{c}{L2P} & \multicolumn{2}{c}{\textsc{TraceLock}} \\
\cmidrule(lr){4-5} \cmidrule(lr){6-7} \cmidrule(lr){8-9} \cmidrule(lr){10-11} \cmidrule(lr){12-13}
\multicolumn{13}{l}{\textbf{Math} (ACC $\uparrow$)} \\
& & & ACC (\%) $\uparrow$ & Avg.\ step $\downarrow$ & ACC (\%) $\uparrow$ & Avg.\ step $\downarrow$ & ACC (\%) $\uparrow$ & Avg.\ step $\downarrow$ & ACC (\%) $\uparrow$ & Avg.\ step $\downarrow$ & ACC (\%) $\uparrow$ & Avg.\ step $\downarrow$ \\
\midrule
\multirow{4}{*}{LLaDA}
& 128 & 32 & 67.1 & 128.0 & 74.4 & 128.0 & 74.8 & 54.3 & 74.8 & \textbf{44.7} & \textbf{74.9} & 84.1 \\
& 128 & 64 & 62.5 & 128.0 & \textbf{73.5} & 128.0 & \textbf{73.5} & 53.4 & 70.5 & \textbf{40.7} & 73.0 & 88.0 \\
& 256 & 64 & 69.1 & 256.0 & 79.8 & 256.0 & 79.4 & 79.4 & 78.3 & \textbf{57.4} & \textbf{80.0} & 144.9 \\
& 256 & 128 & 62.7 & 256.0 & 76.0 & 256.0 & 76.4 & 80.4 & 66.2 & \textbf{60.9} & \textbf{76.5} & 154.3 \\
\midrule
\multirow{4}{*}{Dream}
& 128 & 32 & 59.5 & 128.0 & \textbf{73.8} & 128.0 & 73.2 & \textbf{105.5} & 69.1 & 117.4 & 73.5 & 107.5 \\
& 128 & 64 & 51.2 & 128.0 & 70.0 & 128.0 & 69.5 & 87.6 & 57.3 & 111.2 & \textbf{71.5} & \textbf{58.4} \\
& 256 & 64 & 51.9 & 256.0 & 79.9 & 256.0 & 79.9 & 197.4 & 64.6 & 203.5 & \textbf{80.8} & \textbf{106.1} \\
& 256 & 128 & 43.2 & 256.0 & 59.4 & 256.0 & 58.8 & 143.0 & 24.4 & 139.8 & \textbf{80.7} & \textbf{114.1} \\
\bottomrule
\end{tabular}
}}}

\vspace{0.2em}

\colorbox{QAFill}{%
\parbox{\dimexpr\textwidth-2\fboxsep\relax}{%
\centering
\resizebox{\textwidth}{!}{%
\begin{tabular}{lllcccccccccc}
\toprule
Model & Gen len & Win./Block & \multicolumn{2}{c}{Random} & \multicolumn{2}{c}{Confidence} & \multicolumn{2}{c}{Fast-dLLM} & \multicolumn{2}{c}{L2P} & \multicolumn{2}{c}{\textsc{TraceLock}} \\
\cmidrule(lr){4-5} \cmidrule(lr){6-7} \cmidrule(lr){8-9} \cmidrule(lr){10-11} \cmidrule(lr){12-13}
\multicolumn{13}{l}{\textbf{QA} (average rank $\downarrow$)} \\
& & & Avg.\ rank $\downarrow$ & Avg.\ step $\downarrow$ & Avg.\ rank $\downarrow$ & Avg.\ step $\downarrow$ & Avg.\ rank $\downarrow$ & Avg.\ step $\downarrow$ & Avg.\ rank $\downarrow$ & Avg.\ step $\downarrow$ & Avg.\ rank $\downarrow$ & Avg.\ step $\downarrow$ \\
\midrule
\multirow{4}{*}{LLaDA}
& 128 & 32 & 3.42 & 128.0 & 2.89 & 128.0 & 3.06 & 73.0 & 3.02 & \textbf{64.6} & \textbf{2.60} & 89.7 \\
& 128 & 64 & 3.40 & 128.0 & 2.83 & 128.0 & \textbf{2.81} & 69.6 & 3.10 & \textbf{51.7} & 2.87 & 95.9 \\
& 256 & 64 & 3.51 & 256.0 & 2.74 & 256.0 & 2.87 & 117.3 & 3.21 & \textbf{90.5} & \textbf{2.68} & 160.9 \\
& 256 & 128 & 2.96 & 256.0 & 2.81 & 256.0 & 2.87 & 107.5 & 3.59 & \textbf{60.1} & \textbf{2.76} & 177.1 \\
\midrule
\multirow{4}{*}{Dream}
& 128 & 32 & 3.38 & 128.0 & 2.96 & 128.0 & 3.09 & 106.2 & 3.13 & 109.3 & \textbf{2.31} & \textbf{66.0} \\
& 128 & 64 & 3.12 & 128.0 & 3.04 & 128.0 & 3.19 & 83.5 & 3.23 & 89.2 & \textbf{2.02} & \textbf{69.6} \\
& 256 & 64 & 3.09 & 256.0 & 3.06 & 256.0 & 3.01 & 201.8 & 3.28 & 205.7 & \textbf{2.20} & \textbf{111.4} \\
& 256 & 128 & 2.62 & 256.0 & 3.16 & 256.0 & 3.19 & 140.6 & 3.54 & 133.2 & \textbf{1.73} & \textbf{121.9} \\
\bottomrule
\end{tabular}
}}}

\vspace{0.2em}

\colorbox{CodingFill}{%
\parbox{\dimexpr\textwidth-2\fboxsep\relax}{%
\centering
\resizebox{\textwidth}{!}{%
\begin{tabular}{lllcccccccccc}
\toprule
Model & Gen len & Win./Block & \multicolumn{2}{c}{Random} & \multicolumn{2}{c}{Confidence} & \multicolumn{2}{c}{Fast-dLLM} & \multicolumn{2}{c}{L2P} & \multicolumn{2}{c}{\textsc{TraceLock}} \\
\cmidrule(lr){4-5} \cmidrule(lr){6-7} \cmidrule(lr){8-9} \cmidrule(lr){10-11} \cmidrule(lr){12-13}
\multicolumn{13}{l}{\textbf{Coding} (Pass@1 $\uparrow$)} \\
& & & Pass@1 (\%) $\uparrow$ & Avg.\ step $\downarrow$ & Pass@1 (\%) $\uparrow$ & Avg.\ step $\downarrow$ & Pass@1 (\%) $\uparrow$ & Avg.\ step $\downarrow$ & Pass@1 (\%) $\uparrow$ & Avg.\ step $\downarrow$ & Pass@1 (\%) $\uparrow$ & Avg.\ step $\downarrow$ \\
\midrule
\multirow{4}{*}{LLaDA}
& 128 & 32 & 14.0 & 128.0 & 31.1 & 128.0 & 29.9 & 33.5 & 28.7 & \textbf{29.2} & \textbf{32.3} & 55.5 \\
& 128 & 64 & 11.6 & 128.0 & 31.1 & 128.0 & 28.7 & 33.2 & 24.4 & \textbf{28.0} & \textbf{31.7} & 59.2 \\
& 256 & 64 & 15.2 & 256.0 & 36.0 & 256.0 & 35.4 & 44.9 & 32.3 & \textbf{35.2} & \textbf{37.2} & 86.8 \\
& 256 & 128 & 15.2 & 256.0 & 34.1 & 256.0 & 34.1 & 43.2 & 20.1 & \textbf{37.1} & \textbf{35.4} & 95.7 \\
\midrule
\multirow{4}{*}{Dream}
& 128 & 32 & 22.6 & 128.0 & 47.0 & 128.0 & 45.7 & 100.6 & 48.8 & 106.6 & \textbf{51.2} & \textbf{47.9} \\
& 128 & 64 & 19.5 & 128.0 & 48.8 & 128.0 & 47.6 & 77.9 & 32.9 & 91.0 & \textbf{51.2} & \textbf{47.7} \\
& 256 & 64 & 13.4 & 256.0 & 52.4 & 256.0 & 53.0 & 195.4 & 31.7 & 197.5 & \textbf{56.1} & \textbf{79.9} \\
& 256 & 128 & 15.2 & 256.0 & 44.5 & 256.0 & 45.1 & 134.7 & 4.9 & 131.0 & \textbf{56.1} & \textbf{80.9} \\
\bottomrule
\end{tabular}
}}}
\end{table*}

\paragraph{Additional QA ranking and critic metrics.}
Table~\ref{tab:qa_appendix_top2_critic} reports two complementary QA statistics for the same policy groups and decoding settings as the QA block in Table~\ref{tab:main_results}. The top-two rate is the percentage of evaluated samples whose candidate is placed in the top-two group for the prompt. The critic rate uses the Alpaca reference output as a non-exhaustive reference answer and asks an LLM critic to judge whether the generated answer should be considered correct against that reference. Examples marked not judgeable are skipped in the critic denominator.

\begin{table*}[t]
\centering
\caption{Additional QA metrics on the same settings as the main QA table. Each policy is reported with top-two rate and LLM-critic rate, both in percent. Higher is better for both metrics.}
\label{tab:qa_appendix_top2_critic}
\setlength{\tabcolsep}{3pt}
\renewcommand{\arraystretch}{1.05}
\colorbox{QAFill}{%
\parbox{\dimexpr\textwidth-2\fboxsep\relax}{%
\centering
\resizebox{\textwidth}{!}{%
\begin{tabular}{lllcccccccccc}
\toprule
Model & Gen len & Win./Block & \multicolumn{2}{c}{Random} & \multicolumn{2}{c}{Confidence} & \multicolumn{2}{c}{Fast-dLLM} & \multicolumn{2}{c}{L2P} & \multicolumn{2}{c}{\textsc{TraceLock}} \\
\cmidrule(lr){4-5} \cmidrule(lr){6-7} \cmidrule(lr){8-9} \cmidrule(lr){10-11} \cmidrule(lr){12-13}
\multicolumn{13}{l}{\textbf{QA}} \\
& & & Top-2 $\uparrow$ & Critic $\uparrow$ & Top-2 $\uparrow$ & Critic $\uparrow$ & Top-2 $\uparrow$ & Critic $\uparrow$ & Top-2 $\uparrow$ & Critic $\uparrow$ & Top-2 $\uparrow$ & Critic $\uparrow$ \\
\midrule
\multirow{4}{*}{LLaDA}
& 128 & 32 & 36.0 & 87.0 & 47.0 & 90.4 & 41.0 & 88.9 & 42.5 & 81.2 & \textbf{49.0} & \textbf{91.2} \\
& 128 & 64 & 34.5 & \textbf{85.8} & \textbf{52.6} & 84.8 & 51.0 & 83.8 & 43.0 & 84.3 & 52.0 & 83.9 \\
& 256 & 64 & 31.5 & 84.0 & 57.5 & 81.7 & 51.5 & 85.9 & 35.5 & 85.1 & \textbf{58.0} & \textbf{87.4} \\
& 256 & 128 & 44.0 & 82.4 & 49.5 & 80.5 & 46.0 & 82.8 & 25.1 & 79.1 & \textbf{50.8} & \textbf{85.7} \\
\midrule
\multirow{4}{*}{Dream}
& 128 & 32 & 37.4 & 82.6 & 40.5 & \textbf{84.0} & 33.7 & 80.9 & 33.2 & 83.1 & \textbf{67.3} & 83.5 \\
& 128 & 64 & 45.4 & 82.0 & 37.2 & 77.8 & 31.4 & 84.7 & 27.4 & 83.7 & \textbf{63.4} & \textbf{85.3} \\
& 256 & 64 & 48.4 & 83.1 & 37.5 & 78.6 & 39.2 & 75.0 & 28.1 & \textbf{84.3} & \textbf{69.2} & 83.1 \\
& 256 & 128 & 57.1 & 69.9 & 38.2 & 75.4 & 38.9 & 64.1 & 25.4 & 72.7 & \textbf{63.0} & \textbf{79.3} \\
\bottomrule
\end{tabular}
}}}
\end{table*}

These QA critic numbers are only auxiliary. Alpaca-style prompts include many open-ended requests, and a single reference output is often only one acceptable answer. For stronger correctness claims, automatically checkable settings such as GSM8K math accuracy and HumanEval execution remain more reliable.

\section{Wall-Clock Runtime}

Table~\ref{tab:wall_time} reports average generation wall-clock time in seconds per sample, measured on NVIDIA A40 GPUs. The unstarred \textsc{TraceLock} column is the measured end-to-end runtime of our current PyTorch/Transformers prototype. This prototype runs the controller outside the frozen D-LLM and explicitly materializes intermediate hidden states, which introduces extra GPU-memory traffic and framework overhead that would not necessarily remain in a fused implementation. In a production setting, the controller would ideally be integrated into the D-LLM decoding stack rather than invoked as an external module. In addition, our prototype uses a separate autoencoder projection stack as an engineering device for making hidden features tractable.

To separate this prototype overhead from the cost of the decoding policy itself, we also report \textsc{TraceLock}$^\ast$, a corrected runtime estimate. For each backbone $b$ and generation length $L$, we profile 100 samples with CUDA events and compute
\[
\alpha_{b,L}
=
\frac{
T_{\mathrm{DLM}}^{\mathrm{logits}} + T_{\mathrm{ctrl}}^{\mathrm{core}}
}{
T_{\mathrm{wall}}^{\mathrm{full}}
},
\qquad
T_{\mathrm{TL}^{\ast}}
=
\alpha_{b,L} T_{\mathrm{TL}} .
\]
Here $T_{\mathrm{DLM}}^{\mathrm{logits}}$ is the GPU time of the frozen D-LLM forward without hidden-state extraction, $T_{\mathrm{ctrl}}^{\mathrm{core}}$ is the GPU time of the controller core, and $T_{\mathrm{wall}}^{\mathrm{full}}$ is the measured full prototype runtime. We estimate a separate $\alpha_{b,L}$ for LLaDA and Dream at each generation length, and apply the corresponding factor to the wall-clock values in the table. Thus \textsc{TraceLock}$^\ast$ should be interpreted as an implementation-corrected estimate, while the unstarred column remains the actual measured prototype runtime.

\begin{table*}[t]
\centering
\small
\caption{Average wall-clock generation time in seconds. In each domain block, rows above the midrule are LLaDA results and rows below the midrule are Dream results. \textsc{TraceLock}$^\ast$ applies the correction factor defined above. Lower is better.}
\label{tab:wall_time}
\setlength{\tabcolsep}{2pt}
\renewcommand{\arraystretch}{1.04}
\begin{tabular}{@{}c@{\hspace{0.3em}}c@{\hspace{0.3em}}c@{}}
\colorbox{MathFill}{%
\parbox{0.31\textwidth}{%
\centering
\textbf{Math}\\[0.1em]
\resizebox{\linewidth}{!}{%
\begin{tabular}{llccccc}
\toprule
Gen & Block & Confidence & Fast-dLLM & L2P & \textsc{TraceLock} & \textsc{TraceLock}$^\ast$ \\
\midrule
128 & 32 & 9.10 & 4.14 & \textbf{3.14} & 6.39 & 5.96 \\
128 & 64 & 8.89 & 4.07 & \textbf{2.86} & 6.53 & 6.09 \\
256 & 64 & 24.07 & 8.01 & \textbf{5.28} & 13.97 & 13.45 \\
256 & 128 & 23.31 & 8.10 & \textbf{5.60} & 14.83 & 14.27 \\
\midrule
128 & 32 & 6.41 & 6.16 & 6.02 & 6.69 & \textbf{5.44} \\
128 & 64 & 6.40 & 4.41 & 5.70 & 3.64 & \textbf{2.96} \\
256 & 64 & 19.45 & 14.74 & 15.44 & 9.76 & \textbf{7.76} \\
256 & 128 & 18.96 & 10.70 & 10.63 & 10.49 & \textbf{8.35} \\
\bottomrule
\end{tabular}
}}}
&
\colorbox{QAFill}{%
\parbox{0.31\textwidth}{%
\centering
\textbf{QA}\\[0.1em]
\resizebox{\linewidth}{!}{%
\begin{tabular}{llccccc}
\toprule
Gen & Block & Confidence & Fast-dLLM & L2P & \textsc{TraceLock} & \textsc{TraceLock}$^\ast$ \\
\midrule
128 & 32 & 6.11 & 3.77 & \textbf{3.08} & 4.61 & 4.30 \\
128 & 64 & 6.11 & 3.59 & \textbf{2.47} & 4.94 & 4.61 \\
256 & 64 & 20.67 & 10.37 & \textbf{7.34} & 13.65 & 13.14 \\
256 & 128 & 20.68 & 9.53 & \textbf{4.85} & 15.04 & 14.48 \\
\midrule
128 & 32 & 5.52 & 4.58 & 4.74 & 3.53 & \textbf{2.87} \\
128 & 64 & 5.51 & 3.60 & 3.88 & 3.70 & \textbf{3.01} \\
256 & 64 & 16.46 & 13.02 & 13.27 & 8.76 & \textbf{6.97} \\
256 & 128 & 16.45 & 9.07 & 8.62 & 9.57 & \textbf{7.61} \\
\bottomrule
\end{tabular}
}}}
&
\colorbox{CodingFill}{%
\parbox{0.31\textwidth}{%
\centering
\textbf{Coding}\\[0.1em]
\resizebox{\linewidth}{!}{%
\begin{tabular}{llccccc}
\toprule
Gen & Block & Confidence & Fast-dLLM & L2P & \textsc{TraceLock} & \textsc{TraceLock}$^\ast$ \\
\midrule
128 & 32 & 16.00 & 4.86 & \textbf{4.37} & 7.68 & 7.16 \\
128 & 64 & 16.01 & 4.87 & \textbf{4.14} & 8.27 & 7.71 \\
256 & 64 & 38.56 & 8.05 & \textbf{6.27} & 14.23 & 13.70 \\
256 & 128 & 39.47 & 7.94 & \textbf{6.62} & 15.54 & 14.96 \\
\midrule
128 & 32 & 13.00 & 10.28 & 12.22 & 6.19 & \textbf{5.03} \\
128 & 64 & 13.01 & 8.02 & 10.56 & 6.22 & \textbf{5.05} \\
256 & 64 & 30.87 & 23.67 & 26.71 & 12.15 & \textbf{9.67} \\
256 & 128 & 30.88 & 22.41 & 17.54 & 12.23 & \textbf{9.73} \\
\bottomrule
\end{tabular}
}}}
\end{tabular}
\end{table*}

\section{Feature Details and Ablations}

\paragraph{Detailed feature construction.}
\textsc{TraceLock} uses three late hidden-state snapshots from the frozen D-LLM together with two short-range hidden deltas. Let
\[
h_{t,i}^{(1)}, h_{t,i}^{(2)}, h_{t,i}^{(3)} \in \mathbb{R}^{d}
\]
denote the selected hidden states at position $i$ and step $t$. We define
\[
\Delta_{t,i}^{(1)} = h_{t,i}^{(2)} - h_{t,i}^{(1)}, \qquad
\Delta_{t,i}^{(2)} = h_{t,i}^{(3)} - h_{t,i}^{(2)}.
\]
The hidden states and deltas are compressed before controller training:
\[
z_{t,i}^{(j)} = E_h(h_{t,i}^{(j)}), \qquad
r_{t,i}^{(k)} = E_\Delta(\Delta_{t,i}^{(k)}).
\]
Each token feature is a concatenation of the states, 
\[
q_{t,i} =
\left[
z_{t,i}^{(1)};
z_{t,i}^{(2)};
z_{t,i}^{(3)};
r_{t,i}^{(1)};
r_{t,i}^{(2)}
\right],
\]
followed by a learned projection into the controller space.

\paragraph{Feature ablations.}
Table~\ref{tab:feature_ablation} evaluates hidden-feature variants on held-out decoding traces, measuring future-stability prediction rather than final-answer quality directly.

\begin{table}[h]
\centering
\caption{Feature ablations for future-stability prediction on held-out decoding traces. Metrics are computed on the test split. Best results are bolded.}
\label{tab:feature_ablation}
\begin{tabular}{lccc}
\toprule
Features & AUROC & AP & Acc@0.5 \\
\midrule
Last hidden only & 0.914 & 0.926 & 0.854 \\
Two deltas only & 0.923 & 0.933 & 0.873 \\
Three hidden states only & 0.928 & 0.939 & 0.881 \\
\textbf{Three hidden states + two deltas} & \textbf{0.936} & \textbf{0.945} & \textbf{0.888} \\
\bottomrule
\end{tabular}
\end{table}

Using only the last hidden state already provides a strong stability signal, but it is weaker than feature sets that include multiple hidden snapshots or explicit short-range dynamics. The best results come from combining three hidden-state snapshots with two deltas.

\section{Deployment Mechanism Ablation}

Table~\ref{tab:deployment_ablation} gives the full results for the deployment ablation summarized in Figure~\ref{fig:deployment_ablation_panels}.

\begin{table*}[t]
\centering
\small
\caption{\textsc{TraceLock} deployment ablations across math, QA, and coding. The full model uses both soft-crop and dynamic thresholding. Math reports final-answer accuracy (\%), QA reports average rank, and coding reports Pass@1 (\%). Higher is better for math and coding; lower is better for QA.}
\label{tab:deployment_ablation}
\setlength{\tabcolsep}{4pt}
\renewcommand{\arraystretch}{1.08}
\begin{tabular}{llcccc}
\toprule
Backbone & Gen len & Full & w/o soft-crop & w/o dynamic threshold & w/o both \\
\midrule
\rowcolor{MathFill}
LLaDA & 128 & \textbf{74.9} & 56.5 & 68.7 & 56.2 \\
\rowcolor{MathFill}
LLaDA & 256 & \textbf{80.0} & 58.0 & 70.5 & 55.5 \\
\rowcolor{MathFill}
Dream & 128 & \textbf{73.5} & 70.0 & 71.5 & 69.1 \\
\rowcolor{MathFill}
Dream & 256 & \textbf{80.8} & 77.4 & 77.8 & 76.6 \\
\midrule

\rowcolor{QAFill}
LLaDA & 128 & \textbf{1.81} & 2.96 & 2.18 & 3.06 \\
\rowcolor{QAFill}
LLaDA & 256 & \textbf{1.93} & 2.92 & 2.16 & 2.99 \\
\rowcolor{QAFill}
Dream & 128 & \textbf{2.01} & 2.81 & 2.27 & 2.92 \\
\rowcolor{QAFill}
Dream & 256 & \textbf{2.15} & 2.69 & 2.39 & 2.77 \\
\midrule

\rowcolor{CodingFill}
LLaDA & 128 & \textbf{32.3} & 27.4 & 31.1 & 23.8 \\
\rowcolor{CodingFill}
LLaDA & 256 & \textbf{37.2} & 28.0 & 33.5 & 26.8 \\
\rowcolor{CodingFill}
Dream & 128 & \textbf{51.2} & 47.6 & 49.4 & 45.7 \\
\rowcolor{CodingFill}
Dream & 256 & \textbf{56.1} & 50.0 & 49.4 & 48.8 \\
\bottomrule
\end{tabular}
\end{table*}

\section{Hidden-State Inputs for Blockwise Token Filtering}
\label{sec:hidden_l2p_diagnostic}

The larger-window results suggest that confidence-only token filters can become brittle when the block size increases. To isolate whether this behavior comes from the filter architecture or from the input representation, we run an additional diagnostic experiment that keeps the Learn2PD-style blockwise interface but replaces scalar confidence inputs with hidden-state features. The output interface is therefore the same as Learn2PD, but the input contains richer trace information.

\begin{table}[t]
\centering
\caption{Hidden-state Learn2PD diagnostic on HumanEval at generation length $128$ and block size $64$. Results report Pass@1 (\%). L2P-Hidden keeps the blockwise token-filter interface of Learn2PD, but replaces confidence-only inputs with concatenated hidden-state features. Higher is better.}
\label{tab:hidden_l2p_diagnostic}
\setlength{\tabcolsep}{6pt}
\renewcommand{\arraystretch}{1.08}
\begin{tabular}{lccc}
\toprule
Backbone & L2P-Confidence & L2P-Hidden & \textsc{TraceLock} \\
\midrule
Dream & 32.9 & 46.9 & \textbf{51.2} \\
LLaDA & 24.4 & 28.1 & \textbf{31.7} \\
\bottomrule
\end{tabular}
\end{table}

Table~\ref{tab:hidden_l2p_diagnostic} shows that replacing confidence with hidden-state features substantially improves the Learn2PD-style filter, especially on Dream. \textsc{TraceLock} remains stronger, indicating that representation choice is only part of the design: sequence-conditioned context, trace-state modeling, and adaptive deployment mechanisms also contribute.

\section{Other Variations}
\label{appendix:other-variations}

\subsection{Deployment-Aware Self-Training}
\label{sec:method_self_training}

A controller trained only on heuristic traces faces distribution shift at deployment time. Once \textsc{TraceLock} is inserted into the generation loop, it induces different partially completed sequences, different hidden states, and different future labels. We address this with deployment-aware self-training.

Let $\pi_{\theta_0}$ be a pretrained controller. We decode new prompts with $\pi_{\theta_0}$, record the resulting traces, and again derive labels from the final sequence:
\[
y^{\pi}_{t,i} = \mathbb{I}\left[\hat{x}^{\pi}_{t,i}=x_i^{\pi,\star}\right].
\]
When collecting self-training traces, we use a slightly more conservative acceptance threshold so that the on-policy pool favors higher-quality completed trajectories. After filtering, these on-policy traces are mixed with the original trace pool:
\[
\mathcal{D}_{\text{mix}}
=
\alpha \mathcal{D}_{\text{self}}
+
(1-\alpha)\mathcal{D}_{\text{pre}} .
\]
The old traces are retained to avoid self-reinforcement bias, mode collapse, and forgetting of useful stability patterns learned from the original heuristic distribution.
Table~\ref{tab:tracelock_variants} shows that self-training mainly improves efficiency: \textsc{TraceLock}-ST consistently executes fewer steps than the pretrained controller, while preserving comparable task quality in most settings. We hypothesize that this speedup comes from training on shorter on-policy traces, which teaches the controller to commit earlier under its own deployment distribution.

\begin{table*}[ht]
\centering
\caption{\textsc{TraceLock} adaptation variants on LLaDA. Each cell reports the primary metric and average executed steps as metric / steps. Math uses accuracy, QA uses average rank, and coding uses Pass@1. Higher is better for math and coding; lower is better for QA.}
\label{tab:tracelock_variants}
\setlength{\tabcolsep}{3pt}
\renewcommand{\arraystretch}{1.18}
\begin{tabular}{ccc}
\colorbox{MathFill}{%
\parbox{0.315\textwidth}{%
\centering
\textbf{Math} (ACC / steps)\\[-0em]
\resizebox{\linewidth}{!}{%
\begin{tabular}{lccc}
\toprule
Setting & \textsc{TraceLock} & \textsc{TraceLock}-ST & \textsc{TraceLock}-RL \\
\midrule
128/32 & 74.9 / 84.1 & \textbf{75.8} / \textbf{67.6} & 74.9 / 91.5 \\
128/64 & 73.0 / 88.0 & 73.3 / \textbf{72.4} & 73.3 / 94.8 \\
256/64 & \textbf{80.0} / 144.9 & 78.9 / \textbf{117.9} & 79.4 / 169.2 \\
256/128 & 76.5 / 154.3 & \textbf{77.1} / \textbf{128.9} & 76.2 / 179.1 \\
\bottomrule
\end{tabular}
}}}
&
\colorbox{QAFill}{%
\parbox{0.315\textwidth}{%
\centering
\textbf{QA} (rank / steps)\\[-0em]
\resizebox{\linewidth}{!}{%
\begin{tabular}{lccc}
\toprule
Setting & \textsc{TraceLock} & \textsc{TraceLock}-ST & \textsc{TraceLock}-RL \\
\midrule
128/32 & 2.04 / 89.7 & 2.09 / \textbf{85.1} & \textbf{1.87} / 91.3 \\
128/64 & 2.14 / 95.9 & 2.06 / \textbf{90.2} & \textbf{1.80} / 96.2 \\
256/64 & 2.06 / 160.9 & 2.07 / \textbf{151.8} & \textbf{1.87} / 165.4 \\
256/128 & 2.13 / 177.1 & 2.07 / \textbf{164.6} & \textbf{1.80} / 181.4 \\
\bottomrule
\end{tabular}
}}}
&
\colorbox{CodingFill}{%
\parbox{0.315\textwidth}{%
\centering
\textbf{Coding} (Pass@1 / steps)\\[-0em]
\resizebox{\linewidth}{!}{%
\begin{tabular}{lccc}
\toprule
Setting & \textsc{TraceLock} & \textsc{TraceLock}-ST & \textsc{TraceLock}-RL \\
\midrule
128/32 & \textbf{32.3} / 55.5 & 29.9 / \textbf{44.4} & \textbf{32.3} / 77.9 \\
128/64 & \textbf{31.7} / 59.2 & 31.1 / \textbf{47.6} & 31.1 / 80.7 \\
256/64 & \textbf{37.2} / 86.8 & 32.9 / \textbf{71.4} & 36.6 / 133.0 \\
256/128 & \textbf{35.4} / 95.7 & 33.5 / \textbf{77.6} & 33.5 / 139.5 \\
\bottomrule
\end{tabular}
}}}
\end{tabular}
\end{table*}

\subsection{Beyond Supervised Trace Learning: Reinforcement Learning}
\label{sec:method_rl}

Future-trace supervision is simple and stable, but it is still bounded by the quality of the traces used to train it. To explore whether the policy can move beyond supervised trace imitation, we also implement a reinforcement-learning extension. This component is not required for the main supervised method, but it gives a way to optimize the controller directly against final-answer rewards.

We view the frozen D-LLM and the current partial sequence as the environment. At step $t$, the policy samples token-level accept decisions from Bernoulli probabilities derived from \textsc{TraceLock} scores. To avoid sampling from extremely low-confidence positions, the implementation first applies a sample threshold and can rescale probabilities above that threshold:
\[
\tilde{p}_{t,i}
=
\operatorname{clip}
\left(
\frac{p_{t,i}-\eta_{\text{sample}}}{1-\eta_{\text{sample}}},
\epsilon, 1-\epsilon
\right).
\]
The policy then samples $u_{t,i}\sim \operatorname{Bernoulli}(\tilde{p}_{t,i})$ for eligible positions.

After a full rollout, a reward model scores the completed answer. In our implementation this reward model is Skywork-Reward-V2 \citep{liu2025skywork}. For coding tasks, the reward is adjusted with executable checks. Let $\bar{R}$ be the mean reward in the current group before code-specific adjustment. If a generated code answer fails syntax checking, we subtract $|\bar{R}|$ from its reward; if it passes syntax checking but fails at runtime, we subtract $\frac{1}{2}|\bar{R}|$:  
\[
R_i^{\text{code}}
=
R_i
-
\begin{cases}
|\bar{R}|, & \text{syntax error},\\
\frac{1}{2}|\bar{R}|, & \text{runtime error},\\
0, & \text{otherwise}.
\end{cases}
\]
This prevents obviously invalid code from receiving high preference-model reward by accident, while penalizing syntax errors more strongly than runtime failures. Timeouts are treated as runtime failures.

We optimize the policy with group-relative advantages, following the spirit of GRPO-style training \citep{shao2024deepseekmath}. For a group of $G$ samples with rewards $\{R_j\}_{j=1}^{G}$, the advantage is
\[
A_j = \frac{R_j-\mu_R}{\sigma_R+\epsilon},
\qquad
\mu_R=\frac{1}{G}\sum_j R_j.
\]
The policy loss for a trajectory is
\[
\mathcal{L}_{\text{RL}}
=
-A_j \sum_t \log \pi_\theta(u_t\mid x_t)
-
\beta \sum_t \mathcal{H}\!\left(\pi_\theta(\cdot\mid x_t)\right).
\]
The entropy term is important in practice: it keeps the token-level Bernoulli policy from collapsing too early, and provides a simple exploration pressure while the controller is still uncertain about which acceptance patterns lead to high final reward. 

In practice, this RL stage is sensitive to both data quality and initialization. Training from pretrained or self-trained \textsc{TraceLock} checkpoints is substantially more stable than training the policy from scratch. Without a supervised initialization, the controller starts in a low-reward regime where high-quality trajectories are too sparse, making the reward signal weak and the policy difficult to bootstrap. We therefore treat RL as an extension on top of trace-supervised learning rather than as a standalone replacement: pretraining and self-training provide a strong warm start, and RL then refines the policy beyond what supervised traces alone can provide.

As shown in Table~\ref{tab:tracelock_variants}, RL refinement is most helpful on QA ranking, where it improves average rank across all listed settings. It usually uses more decoding steps, however, and gives mixed results on math and coding. We therefore treat it as a promising adaptation extension rather than part of the default \textsc{TraceLock} configuration.

\begin{figure*}[htbp]
\centering
\setlength{\tabcolsep}{4pt}
\begin{tabular}{ccc}
\textbf{Pre-train} & \textbf{Self-train} & \textbf{RL} \\
\includegraphics[width=0.25\textwidth]{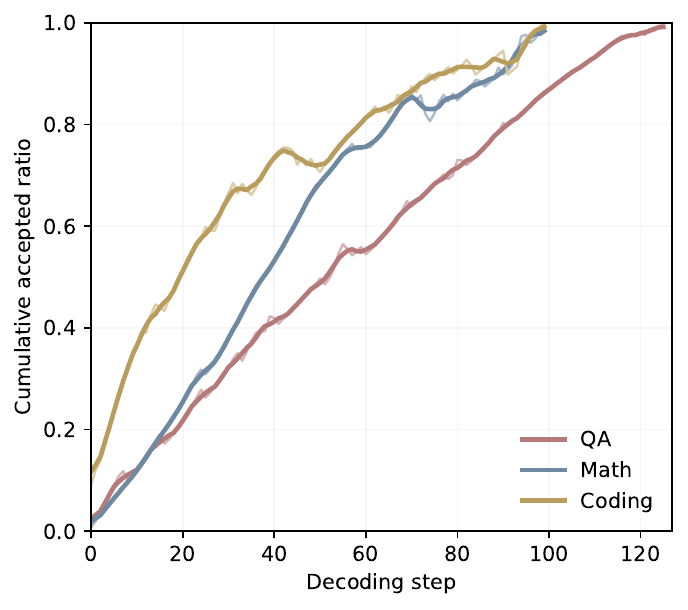} &
\includegraphics[width=0.25\textwidth]{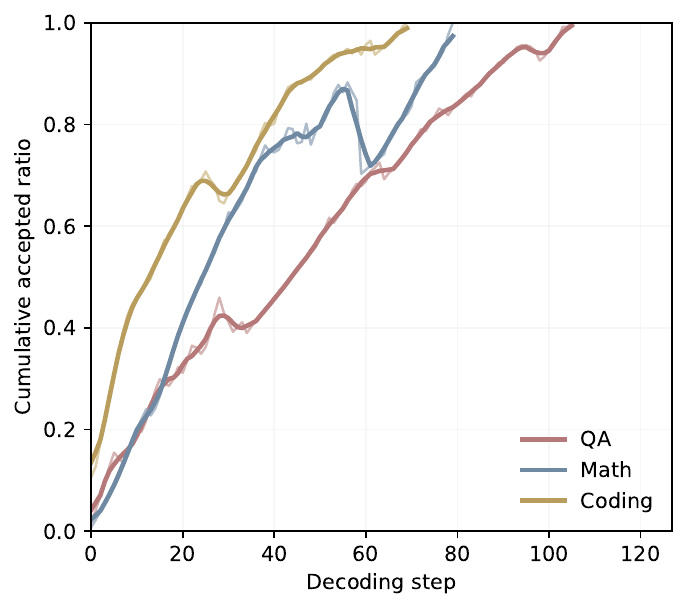} &
\includegraphics[width=0.25\textwidth]{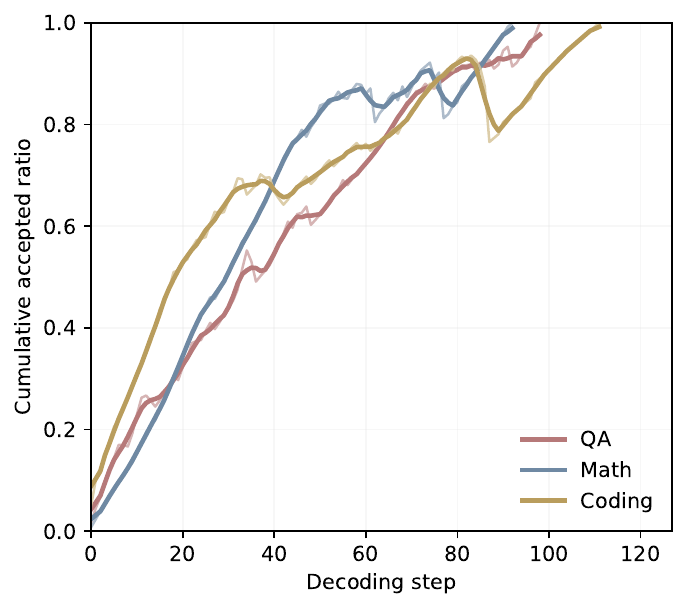} \\
\end{tabular}
\caption{Cumulative acceptance ratio over decoding time on LLaDA with generation length $128$. Left to right, the three panels show the pretrained controller, the self-trained controller, and the RL-refined controller.}
\label{fig:cumulative_acceptance}
\end{figure*}

\begin{figure*}[htbp]
\centering
\setlength{\tabcolsep}{4pt}
\begin{tabular}{ccc}
\textbf{Pre-train} & \textbf{Self-train} & \textbf{RL} \\
\includegraphics[width=0.25\textwidth]{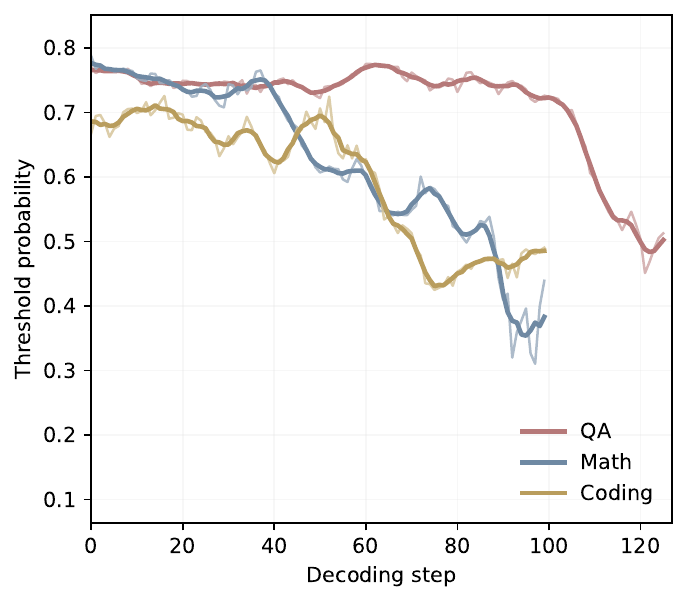} &
\includegraphics[width=0.25\textwidth]{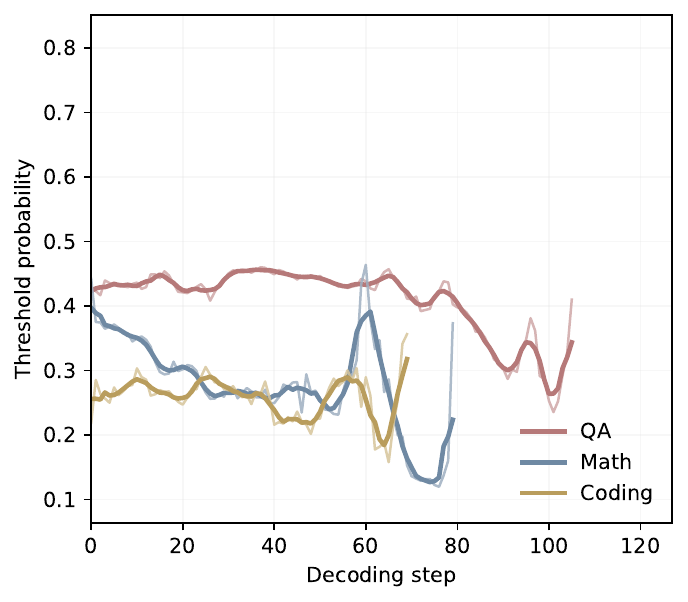} &
\includegraphics[width=0.25\textwidth]{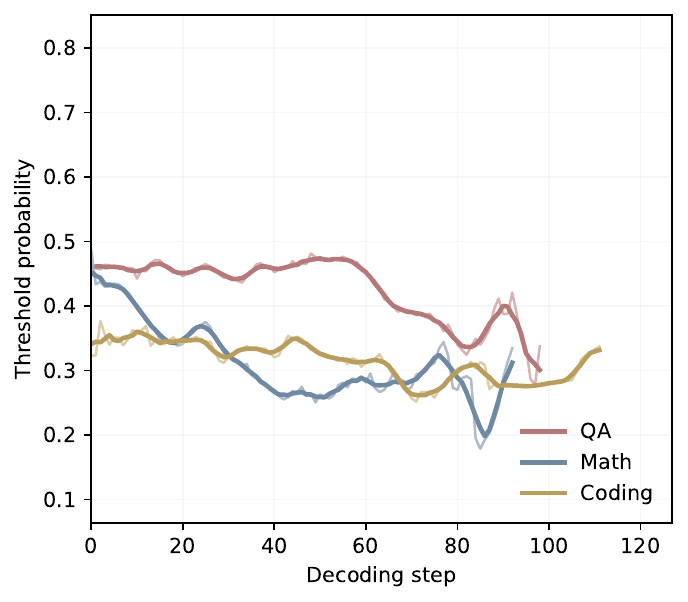} \\
\end{tabular}
\caption{Hidden threshold trajectories over decoding time on LLaDA with generation length $128$. The threshold evolves differently across tasks and controller variants, consistent with sequence-conditioned caution rather than a single global cutoff.}
\label{fig:hidden_threshold}
\end{figure*}

Figures~\ref{fig:cumulative_acceptance} and~\ref{fig:hidden_threshold} summarize decoding dynamics for the three adaptation variants: the former reports the cumulative fraction of accepted tokens over steps, while the latter reports the average learned threshold over steps, with curves aggregated over randomly sampled examples from each task.

\section{Confidence and Trace Diagnostics}
\label{sec:confidence_trace_diagnostics}
\paragraph{Divergence from confidence-based trajectories.}

We compare intermediate trajectories from confidence filtering and \textsc{TraceLock} on the same QA prompts. The diagnostic uses 100 prompts, generation length $128$, block size $128$, and probes both traces every 16 decoding steps. Table~\ref{tab:trace_divergence_summary} reports mask symmetric difference, mask Jaccard overlap, and common-token difference among positions unmasked by both policies. The two policies differ substantially through the middle of decoding, and common-token differences increase at later steps. Figure~\ref{fig:score_confidence} provides a complementary score-level view: \textsc{TraceLock} scores remain positively correlated with confidence, but the correlation changes over decoding time and is far from a fixed confidence ordering.

\begin{table}[t]
\centering
\small
\caption{Trace divergence between confidence filtering and \textsc{TraceLock} on QA. Count metrics are normalized by the generation length of 128.}
\label{tab:trace_divergence_summary}
\setlength{\tabcolsep}{5pt}
\renewcommand{\arraystretch}{1.08}
\begin{tabular}{lccc}
\toprule
Step & Mask sym. diff (\%) & Mask Jaccard (\%) & Common-token diff (\%) \\
\midrule
16 & 36.8 & 54.1 & 1.0 \\
32 & 33.6 & 51.1 & 3.9 \\
48 & 30.6 & 46.9 & 8.3 \\
64 & 25.2 & 45.4 & 17.1 \\
80 & 21.0 & 40.7 & 21.1 \\
96 & 13.1 & 33.4 & 37.2 \\
\bottomrule
\end{tabular}
\end{table}

\begin{figure}[htbp]
\centering
\setlength{\tabcolsep}{4pt}
\begin{tabular}{ccc}
\textbf{QA} & \textbf{Math} & \textbf{Coding} \\
\includegraphics[width=0.25\textwidth]{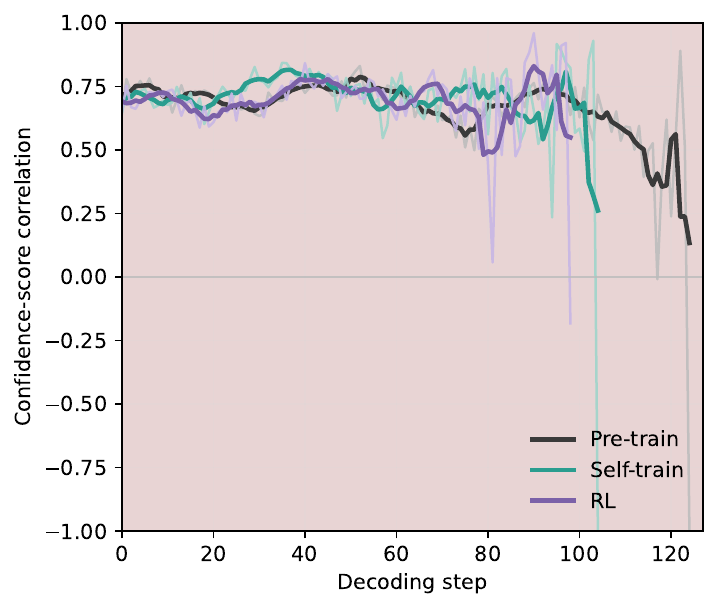} &
\includegraphics[width=0.25\textwidth]{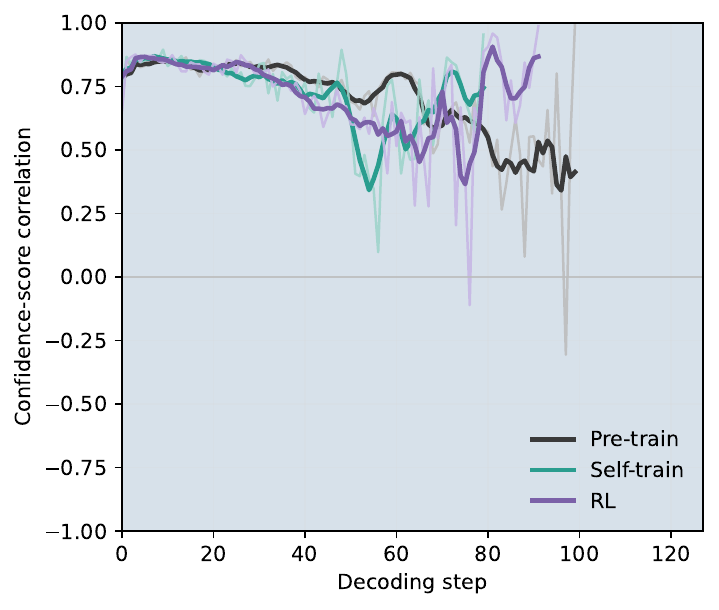} &
\includegraphics[width=0.25\textwidth]{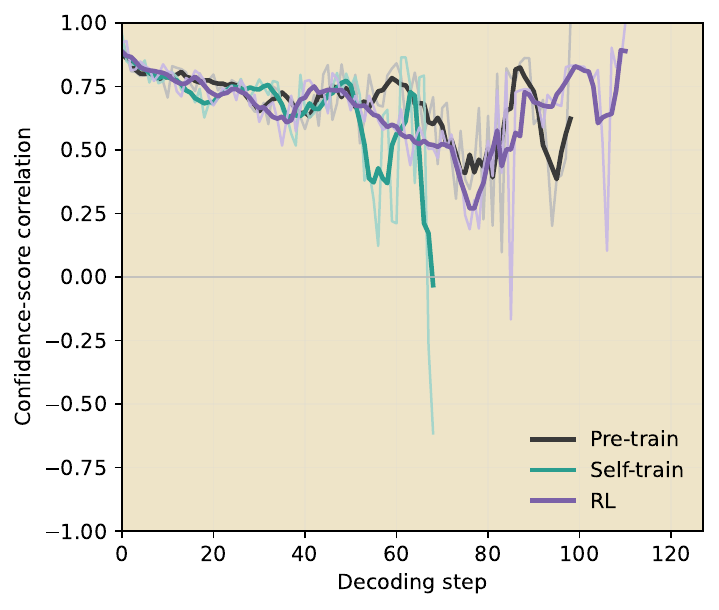} \\
\end{tabular}
\caption{Stepwise correlation between base-model token confidence and \textsc{TraceLock} score on LLaDA with generation length $128$. The learned stability score is positively related to confidence, but the correlation is imperfect and evolves during decoding.}
\label{fig:score_confidence}
\end{figure}

\begin{figure*}[t]
\centering
\includegraphics[width=\textwidth]{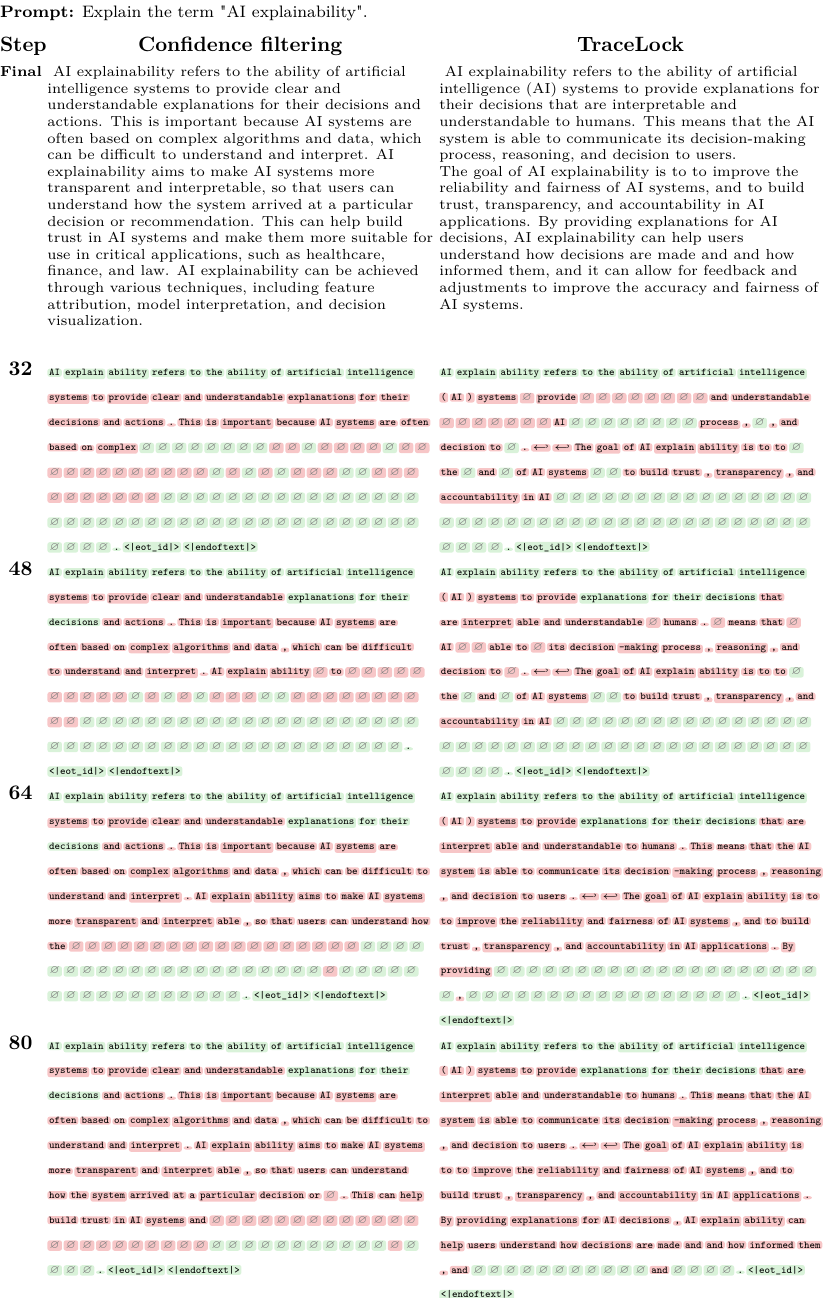}
\caption{Qualitative trace comparison between confidence filtering and \textsc{TraceLock}.}
\label{fig:trace_compare1}
\end{figure*}

\begin{figure*}[t]
\centering
\includegraphics[width=\textwidth]{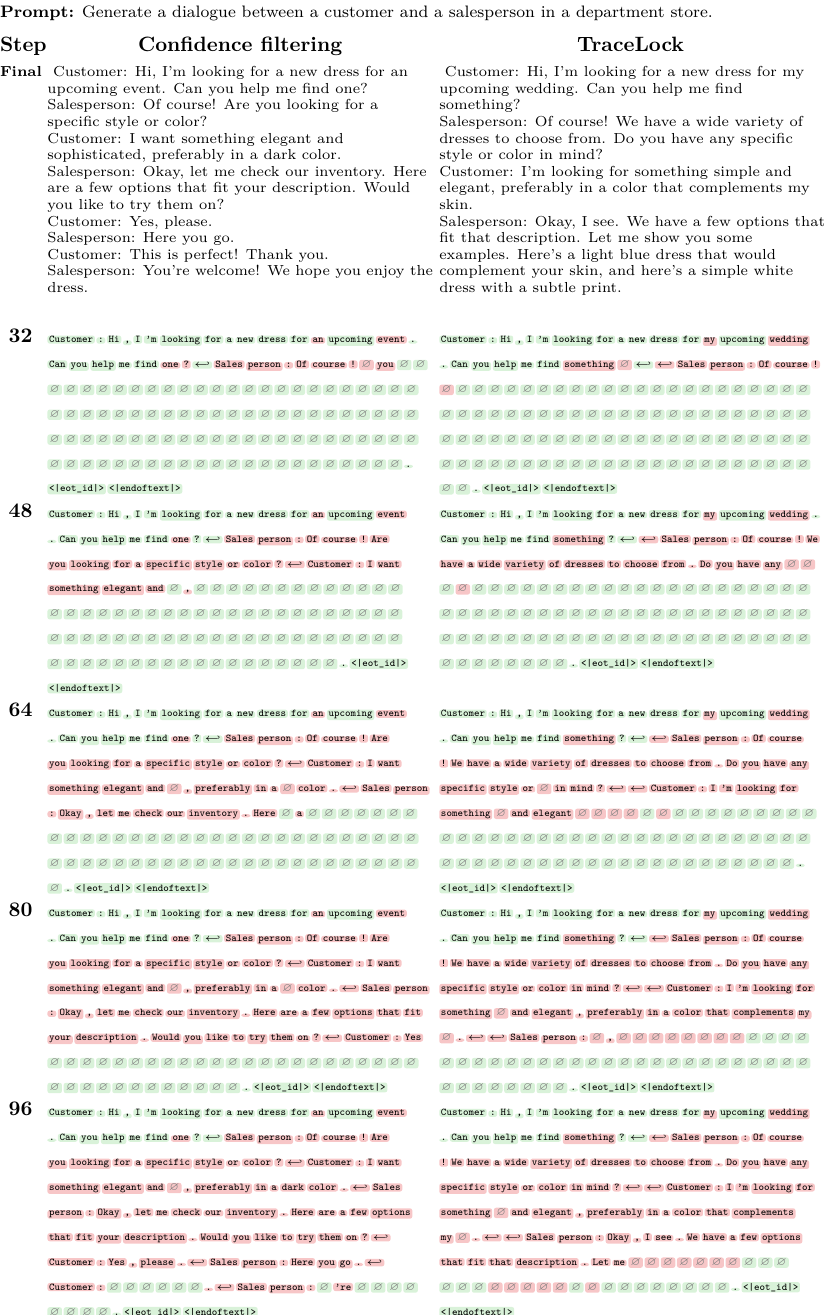}
\caption{Second qualitative trace comparison between confidence filtering and \textsc{TraceLock}.}
\label{fig:trace_compare2}
\end{figure*}

Figures~\ref{fig:trace_compare1} and~\ref{fig:trace_compare2} visualize individual traces. Green cells indicate positions with identical token IDs between the two policies, while red cells indicate mismatched token IDs or a masked/unmasked disagreement.

\section{QA LLM-as-Judge Prompt}
\label{sec:qa_judge_prompt}

For QA evaluation, we use an LLM-as-judge protocol to compare candidate answers generated for the same question. The judge is instructed to score candidates only relative to the other candidates in the same batch, rather than against a global scale across all questions. The resulting scores are then converted into per-question rankings used in the QA evaluation tables. The prompt template is shown below.

\begingroup
\small
\begin{verbatim}
System:
You are a strict scoring assistant. Return exactly one JSON object and nothing else.

User:
Score the candidate answers for the same question.

Rules:
1. Score candidates only relative to each other within this batch, not against a
   global scale across questions.
2. Use a 0 to 10 scale, where higher is better. Decimal scores are allowed.
3. Judge correctness first, then completeness, relevance, reasoning quality,
   factual accuracy, fluency, and clarity.
4. Use the following unified scoring rubric:
   - 9 to 10: The answer is strong and complete. It directly answers the
     question, is factually correct, well-structured, fluent, and covers the
     important aspects with appropriate depth.
   - 7 to 8: The answer is mostly correct and useful, but has some limitations,
     such as missing depth, incomplete coverage of important aspects, minor
     ambiguity, or weaker explanation.
   - 5 to 6: The answer is partially useful but has noticeable problems, such as
     factual mistakes, reasoning gaps, grammar or fluency issues, unclear
     wording, or missing key information.
   - 4 to 5: The answer has major problems. It may contain obvious errors, poor
     structure, substantial omissions, or frequent grammar/clarity issues, while
     still retaining some connection to the question.
   - 2 to 3: The answer is very poor. It may be truncated, self-repeating,
     incoherent, largely irrelevant, or unable to provide a usable response.
   - 0 to 1: The answer is almost unusable or malformed. It does not meaningfully
     answer the question, is nonsensical, empty, or fails to form a coherent
     answer.
5. Do not reward candidate ids, style, verbosity, or politeness by themselves.
6. Penalize hallucinated facts, contradictions, unsupported claims, severe
   repetition, truncation, and answers that fail to address the actual question.
7. The `scores` field must include every candidate exactly once, with no missing
   candidates and no extra names.
8. Equal scores are allowed when candidates are truly tied.
9. Output valid JSON only, matching this schema:
{
  "scores": {
    "candidate_1": 8.4,
    "candidate_2": 6.1
  },
  "reason": "short explanation"
}

Total candidates: {num_candidates}
Candidate ids that MUST all appear exactly once in `scores`:
{candidate_ids}

Question:
{question}

Candidates:
{candidate_answers}
\end{verbatim}
\endgroup

  \section{Experimental Details}
\label{sec:experimental_details}

\paragraph{Trace data.}
For each frozen backbone, we collect 8,000 completed decoding traces for controller training.
The prompts are drawn from the same task families used in our experiments:
mathematical reasoning, open-ended question answering, and code generation.

Each trace is converted into step-level supervision by comparing intermediate token proposals
with the final completed sequence, following the future-stability target in
Section~\ref{sec:method_learning}.
We hold out a validation split of trace samples for controller selection and diagnostic evaluation.
Task-level evaluation uses the datasets and metrics described in Section~\ref{sec:experiments}.

\paragraph{Controller training.}
The main \textsc{TraceLock} controller uses compressed hidden-state features,
a 3-layer Transformer encoder, 8 attention heads, hidden dimension 384,
feed-forward dimension 768, dropout 0.1, position encoding, and a learned dynamic-threshold head.

We train with AdamW using learning rate $10^{-4}$, weight decay 0.01, gradient clipping at 1.0, and random seed 42.
Random prefix cropping is applied over the generated region during training to expose the controller to partial contexts similar to local-window deployment.
Validation is run periodically on held-out trace samples, and the supervised controller checkpoint is selected by validation rollout-proxy $F_{0.5}$.

\paragraph{Compute resources.}
All experiments are run on NVIDIA A40 GPUs.
Collecting the 8,000 supervised training traces for one frozen backbone takes roughly two GPU-hours.
Training the main \textsc{TraceLock} controller runs for about 10,000 optimization steps and takes approximately 20 minutes on one A40.
For the optional adaptation stages, self-training collects about 1,000 on-policy traces and then trains for 1,000 steps, taking about 5 minutes for the training stage.
The reinforcement learning extension uses group size 4 and trains for 1,000 steps, taking about 5 GPU-hours.

\end{document}